\documentclass{mosi}

\usepackage{helvet}
\usepackage{xspace}
\usepackage{enumitem}
\usepackage{environ}
\usepackage{fancyvrb}
\usepackage[T1]{fontenc}
\usepackage{array}
\usepackage{float}
\usepackage{listings}
\usepackage{textcomp}
\usepackage{tcolorbox}
\tcbuselibrary{listings,breakable,skins}

\lstdefinestyle{evaluationprompt}{
  basicstyle=\rmfamily\small\selectfont,
  language={},
  frame=none,
  breaklines=true,
  breakatwhitespace=false,
  breakautoindent=false,
  breakindent=0pt,
  columns=fullflexible,
  keepspaces=true,
  showstringspaces=false,
  upquote=true,
  aboveskip=0pt,
  belowskip=0pt
}
\newcounter{promptlisting}

\newcommand{\suite}{GTAGym\xspace}
\newcommand{\singlefigurewidth}{\columnwidth}
\newcommand{\benchmark}{GTAVBench\xspace}
\newcommand{\benchmarkagi}{GTAVBench-AGI\xspace}

\renewcommand{\suite}{GTAGym\xspace}
\input{templates/openmoss/float-layout}
\graphicspath{{templates/openmoss/}}

\let\OpenMOSSSetAbstract\abstract
\RenewEnviron{abstract}{\global\let\OpenMOSSAbstract\BODY}
\begin{abstract}
AAA games offer a rich testbed for evaluating agent perception, reasoning, and control. However, evaluating whether agents can complete tasks in AAA games is challenging and often human-annotated because task outcomes are often difficult to verify automatically, let alone applying it for Reinforcement Learning (RL).
We introduce \textbf{\suite}, an environment and framework for evaluation and RL on GTA V tasks.
Its evaluation benchmark, \benchmark, derived from GTA V main-story missions, contains 103 online tasks reconstructed across nine task families, alongside 47 image-based shooting and sniping tasks using GTA V imagery.
A web server provides access to the online environment and evaluation interface, supporting parallel testing through screenshots and keyboard--mouse actions with automatic engine-state outcome checks.
We evaluate six frontier LLMs on \benchmark with textual task instructions and on its \benchmarkagi branch, where instructions appear only in the first screenshot and a harness automatically compresses interaction history.
The best configuration achieves 70.9\% overall accuracy in
\benchmarkagi, yet aerial navigation, evasion and interception remain difficult.
\suite also provides a scalable generator that samples task parameters within user-specified ranges and option sets under fixed templates, creating batches of 3D and image-based shooting environments for RL with automatic reward and termination signals.
LoRA RL of Qwen3.8-27B on generated on-foot navigation tasks improves the in-domain performance and transfers to out-of-domain task categories, and native GTA V tasks from 15\% to 23\%.
\end{abstract}

\OpenMOSSSetAbstract{\OpenMOSSAbstract}
\newcommand{\paperabstract}{}

\title{GTAGym: AAA Game Tasks Environment for Evaluation and Reinforcement Learning}
\author{
Jiajun Chen$^{1,2}$,
Shicheng Fang$^{1,2}$,
Jiahao Lu$^{1,2}$,
Yuxin Wang$^{1,2,\dagger}$,
Xipeng Qiu$^{1,2,\dagger}$
\\[2mm]
{\normalfont\normalsize $^{1}$Shanghai Innovation Institute
\qquad $^{2}$Fudan University}
}

\renewcommand{\checkdataformat}[2][]{{{\fontsize{11}{14}\selectfont\textbf{#1:}} {\abstractinfont #2}}}
\checkdata[Code]{\url{https://github.com/ljbro/GTAGym}}
\checkdata[Dataset]{\url{https://huggingface.co/datasets/OpenMOSS-Team/GTAVBench}}

\begin{document}
\maketitle
\begingroup
\renewcommand{\thefootnote}{\fnsymbol{footnote}}
\footnotetext[2]{Corresponding authors:
\href{mailto:wangyuxin@sii.edu.cn}{\texttt{wangyuxin@sii.edu.cn}},
\href{mailto:xpqiu@sii.edu.cn}{\texttt{xpqiu@sii.edu.cn}}.}
\endgroup
% OpenMOSS manuscript snapshot; refresh with scripts/sync_openmoss.py.
\providecommand{\paperabstract}{\begin{abstract}
AAA games offer a rich testbed for evaluating agent perception, reasoning, and control. However, evaluating whether agents can complete tasks in AAA games is challenging and often human-annotated because task outcomes are often difficult to verify automatically, let alone applying it for Reinforcement Learning (RL).
We introduce \textbf{\suite}, an environment and framework for evaluation and RL on GTA V tasks.
Its evaluation benchmark, \benchmark, derived from GTA V main-story missions, contains 103 online tasks reconstructed across nine task families, alongside 47 image-based shooting and sniping tasks using GTA V imagery.
A web server provides access to the online environment and evaluation interface, supporting parallel testing through screenshots and keyboard--mouse actions with automatic engine-state outcome checks.
We evaluate six frontier LLMs on \benchmark with textual task instructions and on its \benchmarkagi branch, where instructions appear only in the first screenshot and a harness automatically compresses interaction history.
The best configuration achieves 70.9\% overall accuracy in
\benchmarkagi, yet aerial navigation, evasion and interception remain difficult.
\suite also provides a scalable generator that samples task parameters within user-specified ranges and option sets under fixed templates, creating batches of 3D and image-based shooting environments for RL with automatic reward and termination signals.
LoRA RL of Qwen3.8-27B on generated on-foot navigation tasks improves the in-domain performance and transfers to out-of-domain task categories, and native GTA V tasks from 15\% to 23\%.
\end{abstract}
}
\paperabstract

\section{Introduction}
\label{sec:introduction}

Building agents that can perceive, reason, and act in real-world environments as flexibly as humans is an ultimate goal of artificial general intelligence (AGI)~\cite{fan2022minedojo}.
Sim-to-real transfer is a common approach to developing real-world agent capabilities: models are trained in simulation and then transferred to physical environments~\cite{tobin2017domainrandomization,peng2018simtoreal}.
Among simulated environments, AAA open-world games offer a rich testbed: navigation, vehicle control, and interactions with moving characters require spatial understanding, temporal reasoning, and action execution~\cite{simateam2024sima,tan2024cradle}.
Vision-language models are increasingly deployed as agents that translate visual observations and natural-language goals into actions~\cite{xie2024osworld,tan2024cradle}.
Evaluating these agents requires measuring whether their actions achieve task goals over an extended interaction.

Existing work follows three broad directions: verifiable tasks in sandbox and purpose-built games~\cite{fan2022minedojo,ouyang2026gameworld,lin2026omnigamearena}, agents that act in commercial 3D worlds~\cite{tan2024cradle,simateam2024sima,tan2025lumine}, and benchmarks that assess gameplay through visual trajectories or offline questions~\cite{zhang2025videogamebench,zhang2026gameverse,wang2026gamehorizon}.
These directions advance task verification, realistic interaction, and broad coverage, respectively.
Bringing them together requires an environment that preserves objectives from real AAA missions, verifies outcomes automatically, and generates new executable tasks for training.
Table~\ref{tab:benchmark_comparison} compares these features across representative benchmarks.

\begin{figure}[t]
    \centering
    \includegraphics[width=0.5\textwidth]{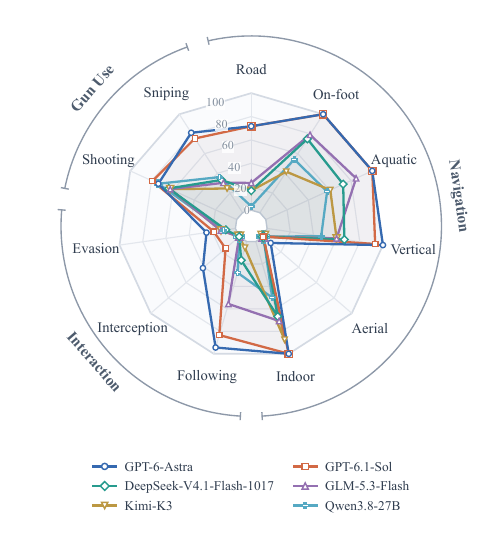}
    \caption{Accuracy (\%) of six models with the Codex harness on \benchmarkagi, across eleven task categories. Outer arcs group tasks into Navigation, Interaction, and Gun Use.}
    \label{fig:performance_radar}
\end{figure}

We introduce \suite, an environment and framework that decomposes main-story missions from GTA V~\cite{rockstar2013gtav} into reusable tasks defined by their goals, interaction stages, completion constraints, and dynamic behaviors.
Its evaluation benchmark, \benchmark, contains 103 online tasks across nine families of navigation, vehicle control, and dynamic interaction, plus 47 image-based shooting and sniping tasks using GTA V imagery.
\benchmark supplies textual task instructions; its \benchmarkagi branch presents them only in the first screenshot, as in-game objectives, and uses a harness with automatic context compression.
A web server exposes a shared screenshot-and-action interface for parallel evaluation, with built-in evaluators checking online outcomes from engine states.
The same framework samples task parameters within user-defined ranges to generate batches of RL environments with automatic reward and termination signals.
Figure~\ref{fig:performance_radar} summarizes GTAVBench-AGI performance with the Codex harness across the eleven evaluation categories.

Our contributions are:
\begin{itemize}
    \item We provide \suite, an environment for parallel evaluation of tasks derived from real AAA gameplay. Built-in evaluators determine task outcomes directly from engine states, removing the need for manual outcome judgments and bringing the programmatic task verification available in Minecraft platforms such as MineDojo~\cite{fan2022minedojo} to tasks derived from GTA V missions.
    \item We construct \benchmark from GTA V main-story missions and introduce its \benchmarkagi branch. The former supplies textual task instructions, while the latter presents goals in the initial game image and uses a harness with automatic context compression. The results reveal persistent weaknesses in completing these mission-derived tasks, particularly in aerial control, evasion and interception.
    \item We equip \suite with a scalable, template-based generator of reinforcement-learning task environments. Each template fixes the scene and task rules while exposing sampling parameters such as movement mode, route length, turn count, and actor speed. Users restrict their ranges or candidate sets to generate 3D and image-based shooting environments in batches, with automatic reward and termination signals to support RL training of exploration, control, and aiming skills.
\end{itemize}

% Draft status, 2026-10-13:
% Task counts follow the current catalog: 104 native 3D + 55 image-based tasks.
% Reconcile counts with the frozen evaluation release before submission.
% Add an empirical findings paragraph after full model and human evaluations.
% Engine states belong to the reconstructed environment, not the original game.
% Current checkpoints use validated action replay, not arbitrary full-state restore.

\section{Related Work}
\label{sec:related_work}

\subsection{Foundation-Model Agents for Video Games}
\label{sec:related_agents}
Foundation models combine visual perception, instruction following, and action selection to operate in commercial games.
Cradle uses screenshots and keyboard--mouse control with planning, memory, and reflection modules, demonstrating main-story progression in Red Dead Redemption 2~\cite{tan2024cradle}.
SIMA learns instruction-following behavior across simulated 3D worlds, using ground-truth checks in research environments and OCR or human judgments in commercial games~\cite{simateam2024sima}; SIMA 2 extends this direction with higher-level reasoning and self-improvement~\cite{simateam2025sima2}.
Lumine trains an end-to-end vision-language policy for keyboard--mouse control in open worlds and uses human evaluation for its task benchmark~\cite{tan2025lumine}.
OctoGTA instruments GTA V through scripting plugins, exposing game entities and programmatic action functions for visually guided task execution~\cite{yang2024octopus}.
\suite instead reconstructs mission-derived tasks with engine-state evaluators and a shared screenshot-and-action interface, supporting parallel evaluation and RL environment generation.

\subsection{Multimodal Game Benchmarks}
\label{sec:related_gamebench}
Games have long served as testbeds for learning and evaluating interactive agents.
Procgen studies RL generalization through procedurally generated levels~\cite{cobbe2020procgen}, while MineDojo provides configurable Minecraft tasks and rewards for open-ended learning~\cite{fan2022minedojo}.
More recent benchmarks evaluate foundation-model agents across diverse games: lmgame-Bench studies perception, memory, and reasoning~\cite{hu2025lmgame}, Orak examines agent modules and fine-tuning across commercial games~\cite{park2025orak}, and VideoGameBench measures visual gameplay progress and the effects of inference latency~\cite{zhang2025videogamebench}.
Specialized benchmarks such as LongPuzzleBench further test long-horizon decision-making in visual puzzles~\cite{zhang2026longpuzzlebench}.

Recent work also develops evaluation protocols for richer game environments.
GameWorld provides state-verifiable browser tasks and parallel execution~\cite{ouyang2026gameworld}, while OmniGameArena introduces custom UE5 games for solo, competitive, and cooperative evaluation~\cite{lin2026omnigamearena}.
For commercial AAA games, GameVerse scores progress through VLM-based milestone matching and studies improvement from video reflection~\cite{zhang2026gameverse}; GameHorizon-Bench evaluates AAA gameplay offline, with its verifiable online track implemented in Minecraft~\cite{wang2026gamehorizon}.
Our work brings objectives extracted from GTA V main-story missions into an environment with built-in outcome checks, parallel web execution, and parameterized RL instance generation.
Table~\ref{tab:benchmark_comparison} compares these features with prior benchmarks.

\begin{table*}[t]
    \centering
    \small
    \setlength{\tabcolsep}{5pt}
    \renewcommand{\arraystretch}{1.18}
    \begin{tabular*}{\textwidth}{@{\extracolsep{\fill}}lccccc@{}}
        \toprule
        Work & \shortstack{3D\\interaction} & \shortstack{AAA open-world\\tasks} & \shortstack{State-based\\scoring} & \shortstack{Web\\execution} & \shortstack{RL instance\\generation} \\
        \midrule
        Procgen~\cite{cobbe2020procgen} & $\times$ & $\times$ & $\checkmark$ & $\times$ & $\checkmark$ \\
        MineDojo~\cite{fan2022minedojo} & $\checkmark$ & $\times$ & $\checkmark$ & $\times$ & $\checkmark$ \\
        VideoGameBench~\cite{zhang2025videogamebench} & $\checkmark$ & $\times$ & $\times$ & $\checkmark$ & $\times$ \\
        GameVerse~\cite{zhang2026gameverse} & $\checkmark$ & $\checkmark$ & $\times$ & $\times$ & $\times$ \\
        GameWorld~\cite{ouyang2026gameworld} & $\checkmark$ & $\times$ & $\checkmark$ & $\checkmark$ & $\times$ \\
        OmniGameArena~\cite{lin2026omnigamearena} & $\checkmark$ & $\times$ & $\checkmark$ & $\times$ & $\times$ \\
        GameHorizon-Bench~\cite{wang2026gamehorizon} & $\checkmark$ & $\times$ & $\times$ & $\times$ & $\times$ \\
        \midrule
        \textbf{\suite (ours)} & $\checkmark$ & $\checkmark$ & $\checkmark$ & $\checkmark$ & $\checkmark$ \\
        \bottomrule
    \end{tabular*}
    \caption{Feature comparison of game benchmarks and their supporting environments. $\checkmark$: explicitly supported in at least one evaluated setting; $\times$: not reported. AAA tasks are interactive objectives from commercial 3D open-world games, including reconstructed missions. State-based scoring is assessed on interactive AAA tasks when present, and on the reported game tasks otherwise. GameVerse uses VLM milestone scoring for its AAA tasks. Web execution includes browser-based subsets; RL generation creates executable training instances from configurations or seeds. Our row covers \suite with \benchmark and \benchmarkagi.}
    \label{tab:benchmark_comparison}
\end{table*}

\section{\suite}
\label{sec:gtabench}

\subsection{\benchmark Task Suite}
\label{sec:benchmark_overview}
\label{sec:task_suite}
\benchmark and its \benchmarkagi branch share a task suite covering eleven dimensions of GTA-style gameplay, spanning navigation, vehicle control, dynamic interaction, and visual aiming.
\suite provides the 3D evaluation environment, spanning urban, indoor, water, and aviation scenarios, with unified task initialization, interaction, and outcome verification.
%The implementation builds on a publicly available city simulation.
Native engine states enable automatic verification of online task outcomes.

\begin{figure*}[t]
    \centering
    \includegraphics[width=\textwidth]{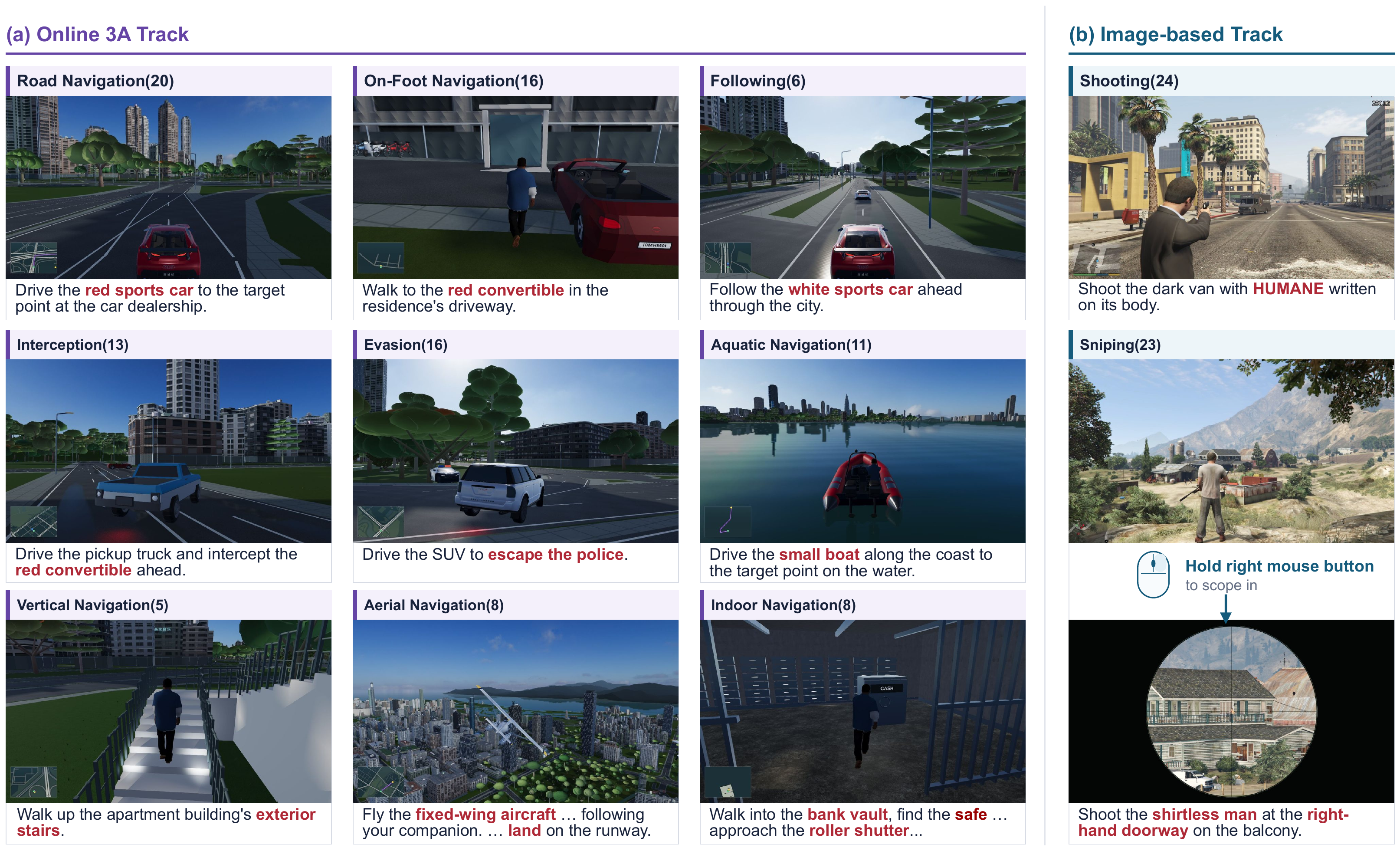}
    \caption{Overview of the \benchmark task suite. (a) The online 3D environments evaluate nine categories of navigation, vehicle control, and dynamic interaction. (b) The image-based environments evaluate shooting and sniping using GTA V imagery. Each category is illustrated with representative visual observations and task instructions. Numbers in parentheses indicate the number of tasks in each category.}
    \label{fig:task_suite}
\end{figure*}

Figure~\ref{fig:task_suite} illustrates the task categories and representative objectives across the two tracks; Table~\ref{tab:task_dimensions} in Appendix~\ref{sec:appendix_task_categories} summarizes their mission requirements.
The online 3D track contains 103 tasks across nine dimensions.
These categories organize the mission-derived goals, interaction stages, completion constraints, and dynamic behaviors that the environment implements.
The image-based environments add 47 shooting and sniping tasks using source gameplay imagery, covering visual target localization and aiming.
Together, the two tracks contain 150 tasks.
This organization supports a capability-level assessment of agents, with results reported separately for the two tracks.
Across both environment types, we group the eleven categories into navigation, interaction, and gun use.

\subsection{Task Construction}
\label{sec:task_construction}
\paragraph{Task extraction and specification.}
We manually decompose 75 GTA V main-story missions into subtasks based on collected walkthrough videos, each corresponding to a distinct stage in a mission.
For each subtask, we specify the goal, required interaction order, completion constraints, and entity behaviors used to reconstruct it in our environment.
These annotations and source game images support the construction of 150 benchmark tasks across eleven categories.

\paragraph{Online scenario reconstruction.}
We build the 3D environment using Babylon.js and implement task controllers that expose task-specific states.
Task specifications configure initial positions, destinations, vehicles, interaction objects, and moving actors.
For navigation tasks, we manually trace paths on the source minimap and seek routes with comparable turn counts in the available city geometry.
We add climbing surfaces and landing areas where required by the task.
Dynamic-interaction tasks use predefined NPC behaviors and mechanics such as collision-induced slowdown.
The resulting scenarios preserve the selected goals, interaction dependencies, and movement constraints while adapting their spatial layout to the evaluation environment.

\paragraph{Image-based task construction.}
Shooting and sniping use a parallel construction process based on source game imagery.
For shooting, we select gameplay frames and associate targets with annotated hit regions.
For sniping, we capture images directly in GTA V to construct distinct unscoped and scoped views with magnification-dependent rendering.

%\paragraph{Mission requirements across categories.}
%The categories specify the mission requirements carried into our evaluation environment.
%Navigation tasks retain destination-reaching objectives and the required movement mode, with indoor and vertical tasks adding object interactions or height constraints.
%Water and flight tasks retain staged objectives, such as boarding a boat before reaching shore, or taking off before completing an aerial objective and landing.
%Following requires maintaining proximity until a moving target reaches its destination; pursuit requires interception before the target arrives or completion of a specified target interaction.
%Evasion requires remaining undetected for a prescribed duration while pursuers search for the agent.

\paragraph{Template instantiation and verification.}
We instantiate each subtask description as a scenario template with an initial state, an instruction, a task budget, and executable success and failure predicates.
The goal specifies the destination, target, or interaction outcome; stage transitions specify the required order of interactions.
Completion predicates check movement mode, height or depth, target distance, and interaction events according to the task.
For following and pursuit, a target controller executes a predefined route that the agent must track through visual observations.
For evasion, pursuer controllers update detection and capture states as the agent moves freely.
Section~\ref{sec:rl_generation} describes how template parameters are sampled to generate RL environments in batches.

% TODO: Document the mapping from source clips to task templates and the
% corresponding goals, stage transitions, constraints, and entity behaviors.
% TODO: Specify and cite the exact GPT model; document route-matching inputs,
% prompts, the turn-count convention, and route validation.

\subsection{Evaluation}
\label{sec:evaluation_protocol}
\label{sec:interaction_interface}

\paragraph{Shared interaction interface.}
\benchmark and \benchmarkagi use the same task instances, action interface, and scoring rules, differing in instruction presentation and context management.
Both branches follow a closed-loop screenshot--action protocol with keyboard--mouse control.
By default, online 3D simulation pauses during inference and advances during action execution.
The agent specifies a hold duration $d\in(0,d_{\max}]$ for each key, with $d_{\max}=1$ second.
Keys execute concurrently, and the turn duration is the longest scheduled hold.
Section~\ref{sec:evaluation_settings} examines alternative duration limits and evaluation without inference-time pausing.
The image-based track uses the same action schema with its aiming interface.
Complete trajectories are recorded for analysis.
The action space is specified in Appendix~\ref{sec:appendix_actions}.

\paragraph{\benchmark.}
Following screenshot-based computer-use evaluation~\cite{xie2024osworld}, agents use a fixed system prompt, a textual task instruction, and screenshot--reply history to generate JSON actions, which an adapter validates and executes.
Online history is capped at ten pairs; exceeding the cap drops the oldest five while preserving the instruction.
Image-based tasks retain the full history.
See Appendix~\ref{sec:appendix_prompts} for prompt and history details.
\paragraph{\benchmarkagi.}
We introduce \benchmarkagi as a separate harness-based evaluation branch of \benchmark.
The task instruction appears only within \textbf{the first screenshot}, presenting the objective through the game view as in real gameplay.
The agent must read the goal from the image and act through the same keyboard--mouse interface.
As interaction history grows, the Codex harness automatically compresses the context supplied to the model~\cite{openai2026codex}.

\paragraph{Scoring.}
Episodes terminate upon task success, task failure, or exhaustion of an applicable turn or game-time budget.
We report category-level accuracy and overall accuracy micro-averaged over 150 tasks.
Task-specific outcome conditions are detailed in Appendix~\ref{sec:appendix_outcomes}; model configurations are specified in Section~\ref{sec:experiments}.

% Counts and scoring follow the October 2026 main results.
% Freeze the context settings with the evaluation release.
% Do not present scripted reference controllers as human baselines.

\subsection{RL Environment Generation}
\label{sec:rl_generation}

\paragraph{Templates and sampling ranges.}
Procedural environment generation provides a way to vary training instances and evaluate RL generalization~\cite{cobbe2020procgen}. To support controlled variation, \suite organizes environment generation around task templates.
 Each template $T$ fixes the scene, available assets, interaction logic, evaluator, and episode budget.
Users control variation within that template through numeric intervals or finite candidate sets $\mathcal{C}$:
\begin{equation}
    \theta\sim P_T(\cdot\mid\mathcal{C}),\qquad
    \mathcal{E}=G(T,\theta;\xi),
\end{equation}
where $\theta$ contains sampled parameter values and $\xi$ is the generation seed.
Parameter definitions, sampling domains, and template-specific settings are provided in Appendix~\ref{sec:appendix_rl_parameters}.

\paragraph{Batch generation for RL.}
The rule-based generator in \suite samples configurations subject to the user's bounds and the template's feasibility constraints.
For example, a driving template can restrict route length to $[200,500]$ meters and turn count to $\{2,3,4,5\}$.
These constraints jointly determine eligible routes; the generator rejects incompatible configurations and reports when no feasible instance can be found.
It constructs the full route and initial state from the accepted values, then saves the resolved configuration and seed for reuse.
Instances share the screenshot--action interface and use task outcomes as sparse rewards, with success and failure ending an episode and budget exhaustion recorded as truncation.
The evaluator uses engine-state predicates for 3D tasks and target-hit checks for image-based tasks.
Repeated sampling generates batches of RL environments within the requested parameter ranges.

\section{Experiments}
\label{sec:experiments}

\subsection{Models and Baselines}
\label{sec:models_baselines}
We evaluate six models on \benchmark under the conventional GUI protocol: GPT-6-Astra~\cite{openai2026gpt6astra}, GPT-6.1-Sol~\cite{openai2026gpt61sol}, DeepSeek-V4.1-Flash-1017~\cite{deepseek2026v41flash}, GLM-5.3-Flash~\cite{zai2026glm53flash}, Kimi-K3~\cite{moonshot2026kimik3}, and Qwen3.8-27B.
We additionally evaluate the \benchmarkagi branch using the Codex harness~\cite{openai2026codex} under the protocol in Section~\ref{sec:evaluation_protocol}.
Additional model configurations are detailed in Appendix~\ref{sec:appendix_model_configurations}.
Navigation and interaction tasks allow up to 300 interaction rounds, while gun-use tasks allow up to 20.
The human baseline evaluation protocol is described in Appendix~\ref{sec:appendix_human_evaluation}.
%
% TODO: Freeze the benchmark and harness versions and document the remaining
% decoding settings.

\subsection{Main Results}
\label{sec:main_results}

Table~\ref{tab:gtabench-main} shows a substantial performance gap between the GPT models and the remaining models under both protocols.
The best model configuration reaches 70.9\% overall accuracy, 18.9 percentage points below the mean human accuracy of 89.8\%.

% Reproduced by scripts/aggregate_multiseed_accuracy.py from three independent seeds.
% Equal seed weights; population standard deviation in percentage points.
% Required packages: \usepackage{booktabs,graphicx}
\begin{table*}[t]
\centering

\begingroup
\fontsize{7.5}{9}\selectfont
\newcommand{\resultstd}[1]{\kern0.5pt{\fontsize{5.5}{6.5}\selectfont\textpm#1}}
\setlength{\tabcolsep}{1.8pt}
\renewcommand{\arraystretch}{1.16}
\resizebox{\textwidth}{!}{%
\begin{tabular}{l*{12}{c}}
\toprule
 & \multicolumn{9}{c}{3D Environments} & \multicolumn{2}{c}{\shortstack{Image-Based\\Environments}} & \\
\cmidrule(lr){2-10}\cmidrule(lr){11-12}
 & \multicolumn{6}{c}{Navigation} & \multicolumn{3}{c}{Interaction} & \multicolumn{2}{c}{Gun Use} & \\
\cmidrule(lr){2-7}\cmidrule(lr){8-10}\cmidrule(lr){11-12}
Model & \shortstack{Road} & \shortstack{On-Foot} & \shortstack{Aquatic} & \shortstack{Vertical} & \shortstack{Aerial} & \shortstack{Indoor} & \shortstack{Following} & \shortstack{Interception} & \shortstack{Evasion} & \shortstack{Shooting} & \shortstack{Sniping} & \textbf{Overall} \\
No. of tasks & 20 & 16 & 11 & 5 & 8 & 8 & 6 & 13 & 16 & 24 & 23 & 150 \\
\midrule
\multicolumn{13}{l}{\textbf{\benchmark}} \\
GPT-6-Astra & 66.7\resultstd{8.5} & 97.9\resultstd{2.9} & 97.0\resultstd{4.3} & \textbf{100.0\resultstd{0.0}} & 0.0\resultstd{0.0} & 95.8\resultstd{5.9} & 83.3\resultstd{0.0} & 5.1\resultstd{3.6} & 18.8\resultstd{5.1} & 76.4\resultstd{3.9} & \textbf{88.4\resultstd{4.1}} & 66.4\resultstd{3.1} \\
GPT-6.1-Sol & 63.3\resultstd{2.4} & 91.7\resultstd{2.9} & 97.0\resultstd{4.3} & \textbf{100.0\resultstd{0.0}} & 0.0\resultstd{0.0} & \textbf{100.0\resultstd{0.0}} & 72.2\resultstd{7.9} & 5.1\resultstd{3.6} & 22.9\resultstd{2.9} & \textbf{81.9\resultstd{2.0}} & 85.5\resultstd{4.1} & 66.0\resultstd{0.9} \\
DeepSeek-V4.1-Flash-1017 & 10.0\resultstd{0.0} & 47.9\resultstd{7.8} & 60.6\resultstd{4.3} & 60.0\resultstd{0.0} & 0.0\resultstd{0.0} & 45.8\resultstd{5.9} & 0.0\resultstd{0.0} & 0.0\resultstd{0.0} & 10.4\resultstd{2.9} & 63.9\resultstd{7.1} & 46.4\resultstd{8.2} & 33.8\resultstd{4.1} \\
GLM-5.3-Flash & 8.3\resultstd{2.4} & 52.1\resultstd{2.9} & 78.8\resultstd{4.3} & 73.3\resultstd{18.9} & 0.0\resultstd{0.0} & 62.5\resultstd{0.0} & 16.7\resultstd{23.6} & 0.0\resultstd{0.0} & 14.6\resultstd{2.9} & 70.8\resultstd{0.0} & 46.4\resultstd{2.0} & 38.9\resultstd{1.3} \\
Kimi-K3 & 11.7\resultstd{4.7} & 41.7\resultstd{2.9} & 54.5\resultstd{0.0} & 33.3\resultstd{18.9} & 0.0\resultstd{0.0} & 45.8\resultstd{5.9} & 0.0\resultstd{0.0} & 0.0\resultstd{0.0} & 10.4\resultstd{2.9} & 55.6\resultstd{2.0} & 31.9\resultstd{2.0} & 28.4\resultstd{0.3} \\
Qwen3.8-27B & 1.7\resultstd{2.4} & 29.2\resultstd{2.9} & 45.5\resultstd{12.9} & 60.0\resultstd{0.0} & 0.0\resultstd{0.0} & 12.5\resultstd{0.0} & 0.0\resultstd{0.0} & 0.0\resultstd{0.0} & 8.3\resultstd{2.9} & 65.3\resultstd{7.9} & 23.2\resultstd{2.0} & 24.2\resultstd{0.8} \\
\midrule
\multicolumn{13}{l}{\textbf{\benchmarkagi(Codex)}} \\
GPT-6-Astra & \textbf{71.7\resultstd{4.7}} & \textbf{100.0\resultstd{0.0}} & \textbf{100.0\resultstd{0.0}} & \textbf{100.0\resultstd{0.0}} & \textbf{8.3\resultstd{5.9}} & \textbf{100.0\resultstd{0.0}} & \textbf{94.4\resultstd{7.9}} & \textbf{41.0\resultstd{18.1}} & \textbf{25.0\resultstd{0.0}} & 73.6\resultstd{2.0} & 81.2\resultstd{8.2} & \textbf{70.9\resultstd{4.4}} \\
GPT-6.1-Sol & \textbf{71.7\resultstd{2.4}} & \textbf{100.0\resultstd{0.0}} & \textbf{100.0\resultstd{0.0}} & 93.3\resultstd{9.4} & 0.0\resultstd{0.0} & \textbf{100.0\resultstd{0.0}} & 83.3\resultstd{0.0} & 15.4\resultstd{0.0} & 18.8\resultstd{0.0} & 79.2\resultstd{5.9} & 75.4\resultstd{2.0} & 66.9\resultstd{1.3} \\
DeepSeek-V4.1-Flash-1017 & 16.7\resultstd{4.7} & 75.0\resultstd{0.0} & 72.7\resultstd{12.9} & 66.7\resultstd{9.4} & 0.0\resultstd{0.0} & 66.7\resultstd{11.8} & 16.7\resultstd{0.0} & 0.0\resultstd{0.0} & 8.3\resultstd{5.9} & 63.9\resultstd{7.9} & 33.3\resultstd{8.2} & 38.2\resultstd{2.5} \\
GLM-5.3-Flash & 23.3\resultstd{2.4} & 79.2\resultstd{2.9} & 84.8\resultstd{4.3} & 60.0\resultstd{28.3} & 0.0\resultstd{0.0} & 70.8\resultstd{5.9} & 55.6\resultstd{15.7} & 0.0\resultstd{0.0} & 12.5\resultstd{0.0} & 62.5\resultstd{5.9} & 30.4\resultstd{0.0} & 41.8\resultstd{1.3} \\
Kimi-K3 & 16.7\resultstd{4.7} & 41.7\resultstd{2.9} & 60.6\resultstd{4.3} & 60.0\resultstd{0.0} & 0.0\resultstd{0.0} & 87.5\resultstd{17.7} & 5.6\resultstd{7.9} & 0.0\resultstd{0.0} & 12.5\resultstd{0.0} & 62.5\resultstd{5.9} & 24.6\resultstd{8.2} & 33.1\resultstd{2.2} \\
Qwen3.8-27B & 3.3\resultstd{2.4} & 54.2\resultstd{5.9} & 57.6\resultstd{17.1} & 46.7\resultstd{9.4} & 0.0\resultstd{0.0} & 50.0\resultstd{0.0} & 27.8\resultstd{7.9} & 0.0\resultstd{0.0} & 10.4\resultstd{5.9} & 73.6\resultstd{2.0} & 36.2\resultstd{8.2} & 34.2\resultstd{3.1} \\
\midrule
% Human baseline: equal participant weights; Overall is task-weighted accuracy.
Human & 90.0\resultstd{4.1} & 100.0\resultstd{0.0} & 100.0\resultstd{0.0} & 100.0\resultstd{0.0} & 45.8\resultstd{5.9} & 100.0\resultstd{0.0} & 100.0\resultstd{0.0} & 82.1\resultstd{3.6} & 81.3\resultstd{5.1} & 88.9\resultstd{5.2} & 95.7\resultstd{3.5} & 89.8\resultstd{1.4} \\
\bottomrule
\end{tabular}%
}
\caption{Task-level accuracy (\%) on \benchmark and \benchmarkagi.
The 3D environments contain 103 tasks in nine categories; the image-based environments contain 47 tasks in two gun-use categories.
Entries show means and population standard deviations.
Bold denotes the best nonzero model score in each column, including ties.}
\label{tab:gtabench-main}
\endgroup
\end{table*}

\paragraph{\benchmark and \benchmarkagi.}
Compared with \benchmark, the \benchmarkagi protocol increases overall accuracy for all six models by 0.9--10.0 percentage points.
The gains are most consistent in navigation, while shooting and sniping show mixed changes across models.

\paragraph{Dynamic control remains a bottleneck.}
Both GPT models reliably complete on-foot, indoor, and vertical navigation, yet struggle when success requires sustained regulation of vehicle motion or an evolving target relationship.
Interception particularly separates Astra from the other models, while aerial navigation and evasion remain difficult across the evaluated configurations.
To understand these differences, we examine task-specific failure modes in Section~\ref{sec:category_control}.

\begin{figure}[tb]
    \centering
    \includegraphics[width=\singlefigurewidth]{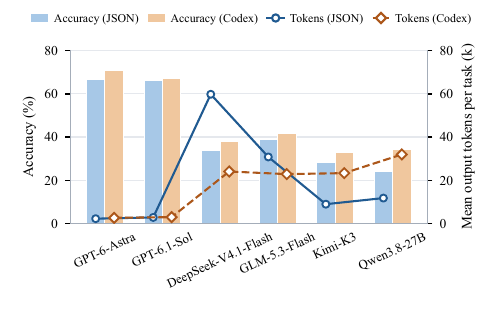}
    \caption{Accuracy and mean output tokens per task using the main-table configurations.}
    \label{fig:performance_tokens}
\end{figure}

\paragraph{Performance and output-token usage.}
Figure~\ref{fig:performance_tokens} compares task accuracy and mean output tokens per task across models and evaluation protocols.
GPT models retain the strongest accuracy--token tradeoff: Astra without the harness uses the fewest output tokens, while Astra with Codex achieves the highest accuracy.
The other models spend substantially more tokens without matching GPT's task success; DeepSeek, for example, uses approximately $26\times$ as many output tokens as Astra under the JSON protocol. Figure~\ref{fig:reasoning_effort} in Appendix compares accuracy and token usage across reasoning-effort settings.

\subsection{Reinforcement Learning}
\label{sec:lora_rl_27b_results}
We study whether reinforcement learning with low-rank adaptation (LoRA)~\cite{hu2022lora} on generated on-foot environments improves task completion and transfers to other categories.

\paragraph{Training environment generation.}
Using the generator in Section~\ref{sec:rl_generation}, we create 50 on-foot navigation instances, matching the benchmark category's joint route-length and turn-count distribution and adding destination landmarks to some instances.
The generated routes are spatially separated from the benchmark's on-foot routes; all benchmark tasks remain excluded from training.
Generation and optimization details are provided in Appendix~\ref{sec:appendix_lora_rl}.

\paragraph{Training.}
We train Qwen3.8-27B using Group Relative Policy Optimization (GRPO)~\cite{shao2024deepseekmath} inside the Codex harness.
LoRA uses rank 32 and $\alpha=64$ on the attention query and value projections, with a learning rate of $10^{-6}$.
Each update samples eight rollouts for each of four tasks.
For a rollout $\tau$, we assign the terminal reward
\begin{equation}
    \begin{aligned}
        R(\tau) &= S+0.3(1-S)P(\tau)-0.1\cdot\mathbf{1}_{\{N>100\}}, \\
        P(\tau) &= \operatorname{clip}\!\left(1-\frac{d_T}{d_0},0,1\right).
    \end{aligned}
    \label{eq:rl_terminal_reward}
\end{equation}
where $S\in\{0,1\}$ indicates task success, $d_0>0$ and $d_T$ are the initial and final traversable distances to the goal, and $N$ counts interaction rounds.
Distance progress rewards unsuccessful rollouts, while exceeding 100 rounds incurs a single $0.1$ deduction.
Distance estimation and scoring details appear in Appendix~\ref{sec:appendix_rl_reward}.

\begin{figure}[tb]
    \centering
    \includegraphics[width=0.5\textwidth]{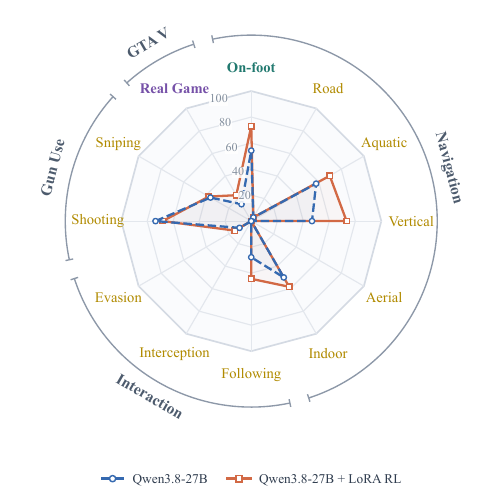}
    \caption{Accuracy (\%) before and after LoRA RL. Green marks the ID category (on-foot navigation), yellow marks OOD categories, and the purple Real Game axis aggregates 100 native GTA V tasks. }
    \label{fig:rl-transfer}
\end{figure}

\paragraph{Results.}
Figure~\ref{fig:rl-transfer} compares mean accuracy across three seeds before and after RL training.
Accuracy rises from 54.2\% to 72.9\% on held-out on-foot tasks and from 34.2\% to 38.9\% overall.
Across OOD categories, accuracy rises from 31.8\% to 34.8\%, with gains in vertical navigation (46.7\% to 73.3\%) and aquatic navigation (57.6\% to 69.7\%).
Road driving remains at 3.3\%, aerial navigation and interception remain at 0\%, and shooting accuracy drops from 73.6\% to 68.1\%.
In blocked trajectories, the base model often repeats ineffective actions, while the trained model uses action feedback to switch strategies and resume progress.

\paragraph{Transfer to native GTA V.}
We compare the base and RL-trained Qwen3.8-27B on 100 native GTA V navigation and interaction tasks through the same screenshot--action interface and Codex harness.
Recorded success rises from 15\% to 23\%, as shown by the Real Game axis in Figure~\ref{fig:rl-transfer}.
Source missions and video-based human verification are detailed in Appendix~\ref{sec:appendix_native_gta_transfer}.

\section{Analysis}
\label{sec:analysis}

We analyze failure modes, error recovery, harness trade-offs, and evaluation settings.
% First, we examine failure modes and recovery along task trajectories.
% Second, we study goal retention and game behavior under different context-management settings.
% Third, we examine the effects of action-duration limits and inference-time pausing.

% Declare the evaluation tables early so the wide ablation table can float
% onto the same page as Section 5.3, while preserving table numbering.
%\input{tables/evaluation-efficiency}

\subsection{Failure Mode Analysis}
\label{sec:category_control}
Trajectory inspection reveals four recurring failure mechanisms. We contrast road-navigation behavior across models, then examine dynamic-control failures that persist even in frontier models. Table~\ref{tab:dynamic_error_cases} in the appendix provides paired observations and executed actions for all four mechanisms.

\noindent\textbf{Mistimed steering.}
In road navigation, non-GPT models sometimes recognize an upcoming turn on the minimap but initiate steering too late or hold it too long for the current speed, overshooting the road. GPT models more effectively translate these cues into early braking and smaller heading corrections.
% The gap therefore extends beyond route recognition to timely, speed-aware action execution.

\noindent\textbf{Late descent correction.}
In aerial navigation, reaching the runway does not ensure a safe landing. In the illustrated GPT trajectory, reducing throttle fails to arrest a rapid descent; pitch correction follows too late to prevent a crash. Coordinating attitude, speed, and descent remains difficult even when the destination is correctly identified.

\noindent\textbf{Failure to sustain blocking.}
In interception, slowing the target through collision is only an intermediate step. The agent continues accelerating after impact and loses contact instead of braking to hold a blocking position, allowing the target to resume its route.

\noindent\textbf{Wall entrapment.}
In evasion, the agent continues applying forward throttle and steering against a wall after hitting a building, producing no displacement before eventually reversing. Such delays interrupt escape progress.

Together, these cases expose difficulty anticipating motion and switching control strategies as the vehicle, target, or obstacle relationship changes.

\subsection{Error Recovery Across Models}
\label{sec:failure_modes}
To study the recovery ability of different models, we manually identify and annotate the first task-relevant error in each trajectory.
We group first errors into four capability categories: Perception, Planning, Calibration, and Regulation.
Across 900 trajectories from six models, our analysis includes 565 with an identified and categorized first error.
Figure~\ref{fig:failure_capability_heatmap} shows that Calibration and Regulation dominate, together accounting for 63.2\% of assigned first errors.
This concentration suggests a bottleneck in closed-loop control: calibrating action direction, magnitude, and duration, and adjusting maneuvers as motion and target states change. We then examine whether the agent subsequently recovers.
Figure~\ref{fig:failure_recovery_heatmap} shows that GPT models recover more reliably: Astra and Sol restore task progress within ten subsequent decisions in 65.7\% and 49.3\% of categorized first-error cases, compared with 21.6\%--25.6\% for the other models.
In road-navigation trajectories, GPT models correct off-road deviations and rejoin the intended route, resuming progress after local control errors.

\begin{figure*}[t]
    \begin{minipage}[t]{0.48\textwidth}
        \input{figures/failure-heatmap}
    \end{minipage}\hfill
    \begin{minipage}[t]{0.48\textwidth}
        \input{figures/failure-recovery-heatmap}
    \end{minipage}
\end{figure*}

\subsection{Harness Trade-offs}
\label{sec:harness_context}
We study the instruction-presentation and context-management differences between \benchmark and \benchmarkagi.
Compared with the JSON runner's ten-image window, the Codex harness preserves longer histories and uses textual summaries to maintain continuity.
This provides memory of attempted strategies, explored rooms, and interaction feedback, helping agents revise their plans and avoid repeated ineffective actions.
In motorcycle interception, GPT-6-Astra recalls a blocked approach and changes its maneuver before re-engaging; in indoor navigation, DeepSeek-Flash retains explored rooms and unsuccessful door interactions to guide further search.
However, longer histories and compression can also preserve mistaken interpretations and weaken goal adherence.
Kimi-K3 initially identifies a rooftop objective correctly but, after compression, substitutes a driving objective and later drops the original goal from its summary.
GLM-5.3-Flash similarly carries reversed minimap-marker identities into subsequent decisions.
These cases illustrate trajectory drift and goal loss around context handoffs.

\subsection{Effects of Evaluation Settings}
\label{sec:evaluation_settings}
\begin{figure*}[t]
    \centering
    \includegraphics[width=\textwidth]{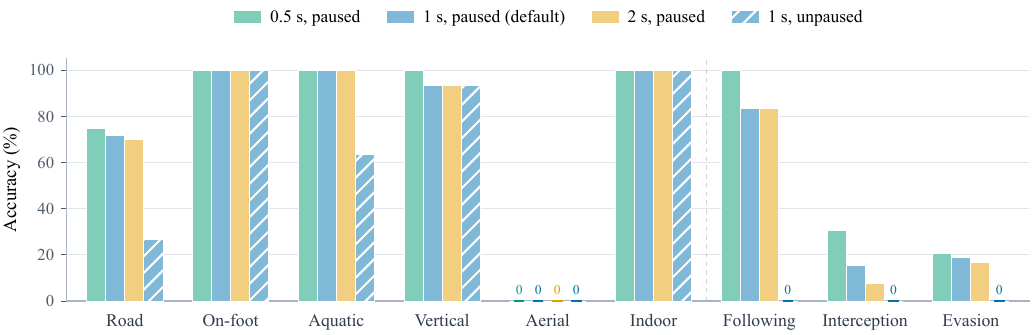}
    \caption{Accuracy (\%) of GPT-6.1-Sol with Codex across nine 3D task categories under four action-duration and inference-time pausing settings.}
    \label{fig:evaluation_settings_radar}
\end{figure*}

We examine how the timing settings of the online 3D evaluation protocol affect task completion.
Both ablations use GPT-6.1-Sol with the Codex harness on the 103 online 3D tasks in \benchmarkagi.
Figure~\ref{fig:evaluation_settings_radar} compares the four distinct settings across the two ablations.
\paragraph{Maximum action duration.}
We compare three values of $d_{\max}$: 0.5\,s, 1\,s (default), and 2\,s.
The agent chooses each key's duration within $(0,d_{\max}]$; simulation pauses during inference in all three conditions.
Relative to the 1\,s baseline (62.1\%), reducing the cap to 0.5\,s raises overall accuracy to 66.3\%, while increasing it to 2\,s lowers accuracy to 60.5\%.
Figure~\ref{fig:evaluation_settings_radar} shows the largest gains in following and interception, where shorter actions provide more frequent visual feedback.
The tradeoff is more interaction and longer cumulative context: reducing the cap from 1\,s to 0.5\,s increases mean rounds from 62.6 to 121.2, input tokens from 4.32M to 11.20M, and output tokens from 4.27k to 7.66k. The mean maximum duration selected within an action increases with the allowed cap: 0.494, 0.916, and 1.468\,s.
At the 2\,s cap, 66.6\% of actions exceed 1\,s, although shorter inputs remain available. This suggests that model does not reliably choose appropriate action durations under our interface.

\paragraph{Evaluation without pausing.}
With $d_{\max}=1$ second, we let the simulation continue during model inference, releasing each key after its specified duration.
Without pausing, overall accuracy drops by 22.3 percentage points to 39.8\%.
Following, interception, and police evasion fall to zero success, while on-foot and indoor navigation retain 100\% accuracy and vertical navigation remains at 93.3\% (Figure~\ref{fig:evaluation_settings_radar}).
This contrast points to inference latency as a bottleneck for dynamic interaction: after a short action ends, the environment continues evolving while the agent waits for its next decision.
Targets can move beyond following range or reach their destinations, and police can capture the player before another corrective action is available.
Tasks that tolerate waiting remain more tractable.
Thus, real-time performance depends on whether the inference-and-action cycle is fast enough for the task's response window.

\section{Conclusion}
\label{sec:conclusion}

We presented \suite, a web-server-based environment and framework for evaluation and RL on tasks derived from GTA V main-story missions.
\benchmark provides text-instructed evaluation, while its \benchmarkagi branch presents task goals only in the first screenshot and uses a harness with automatic context compression.
Both support parallel task execution and automatic outcome verification through a unified interface.
Experiments show that current agents remain limited in aerial control, evasion and interception.
The environment generator in \suite complements \benchmark by sampling numeric and categorical parameters within user-defined bounds under fixed templates, creating 3D and image-based shooting tasks with reward and termination signals for RL training.
LoRA RL on generated on-foot navigation tasks improves Qwen3.8-27B's task completion, with gains extending to unseen task categories and native GTA V missions.
Together, these components provide a basis for evaluating and improving agents on tasks derived from real AAA gameplay.

\section*{Acknowledgements}
We thank the developers and contributors of \href{https://github.com/linranff/GTA_SZ}{\texttt{GTA\_SZ}} for making their game environment openly available.
Our reconstructed 3D environment builds on their work.

\FloatBarrier
\clearpage
{\small
\bibliographystyle{ieeenat_fullname}
\bibliography{references}
}
\clearpage
\appendix
\section{Evaluation Details}
\label{sec:appendix_evaluation}

\subsection{Task Categories and Mission Requirements}
\label{sec:appendix_task_categories}
Table~\ref{tab:task_dimensions} summarizes the eleven task categories and their mission requirements.

\begin{table}[H]
    \centering
    \small
    \setlength{\tabcolsep}{3pt}
    \begin{tabular}{@{}p{0.36\columnwidth}p{0.49\columnwidth}@{}}
        \hline
        Category & Mission requirements \\
        \hline
        \multicolumn{2}{@{}l}{\textit{Online 3D track: Navigation}} \\
        Road & Reach a destination in a road vehicle \\
        Aerial & Takeoff, aerial objectives, and landing \\
        Aquatic & Depth, boarding, and shore-arrival objectives \\
        Indoor & Reach interior goals and interact with doors or objects \\
        On-Foot & Reach a destination on foot \\
        Vertical & Reach a target height using stairs or climbing \\
        \multicolumn{2}{@{}l}{\textit{Online 3D track: Interaction}} \\
        Following & Maintain proximity until the target arrives \\
        Interception & Stop or disable a moving target \\
        Evasion & Avoid detection and capture for a required duration \\
        \hline
        \multicolumn{2}{@{}l}{\textit{Image-based track: Gun Use}} \\
        Shooting & Locate and hit a target in gameplay imagery \\
        Sniping & Use scope and zoom to hit a target or clear all specified targets \\
        \hline
    \end{tabular}
    \caption{Task categories and the mission requirements implemented in the benchmark. Each category groups objectives with shared interaction and completion requirements.}
    \label{tab:task_dimensions}
\end{table}

\subsection{Task-Specific Outcome Conditions}
\label{sec:appendix_outcomes}
The evaluator determines completion from engine states and task events for online tasks, and from annotated hit regions for image-based tasks.
Success is binary: incomplete route progress, unfinished interactions, and partially cleared target sets receive no partial credit.
Both evaluation branches use the same predicates; an agent's declaration of completion cannot substitute for satisfying them.

\paragraph{Road and On-Foot.}
The agent must enter the destination region using the required movement mode: driving for road tasks and walking for on-foot tasks.
Default horizontal tolerances are 24\,m and 3\,m, respectively, with smaller task-specific regions for precise destinations.
Arrival tests include the segment traversed between consecutive physics updates, so a fast vehicle crossing the region is not missed between frames.
Walking arrivals require ground support and cannot be triggered during a jump or traversal animation.
The displayed route and intermediate turns guide navigation without imposing mandatory checkpoints; physically feasible shortcuts are allowed.
Vehicle-acquisition variants additionally require the agent to find and enter a vehicle. Their ground-truth reference sequence includes the initial walking and boarding stages.
Driving into deep water initiates irreversible sinking and prevents arrival success.

\paragraph{Vertical and Indoor.}
These templates add height and interaction requirements to arrival.
For example, the pickup-to-rooftop task requires a grounded walking agent within 1\,m of the goal and 0.15\,m of the roof height; the ladder-to-rooftop task uses 1.1\,m horizontally and 0.35\,m vertically.
Passing beneath an elevated goal does not complete it.
Interior exits can use an oriented passage region instead of a small circular goal, and the agent must reach the correct floor.
Required interactions are checked through explicit mission state: the bank task requires pressing E within 1.5\,m of the cash interaction point, opening the exit shutter, and leaving the building.
The cover task requires pressing Q between the cover and the rear wall, within 0.85\,m of its interaction point, and completing the crouched pose.
Other interiors can require an outward-and-return sequence before the final exit becomes valid.

\paragraph{Following.}
The target follows its configured route while the evaluator checks the agent--target distance throughout the episode.
The maximum horizontal separation is 20\,m on foot and 200\,m in road vehicles.
Exceeding the applicable threshold causes immediate failure, whether the agent falls behind or moves too far ahead.
Success occurs only after the target reaches its destination and the agent remains in the required movement state.
Thus the agent must sustain following throughout the NPC's traversal.
Returning within range after a violation does not reverse a terminal failure.

\paragraph{Interception.}
For ordinary vehicle targets, the agent must establish frontal collision contact and continue blocking the target.
Success requires the target's speed to remain at or below 3\,m/s for 1.5\,s while the agent remains ahead, with a collision gap below 2.5\,m and its own speed at or below 4\,m/s.
Braking from a collision is integrated over time; a naturally slow corner does not satisfy the interception condition.
Motorcycle targets instead require three distinct knockdowns, each caused by an impact with at least 2\,m/s closing speed.
Further hits count only after recovery and separation; the third knockdown completes after the visible fall is complete and the motorcycle slows to at most 1\,m/s.
A companion-recovery variant uses a separate predicate: the agent must align alongside the moving truck and complete the companion transfer.
For these tasks, the target reaching its endpoint before completion causes failure, including ties within a physics update.

\paragraph{Evasion.}
Evasion has no destination or prescribed route.
Evasion has a fixed maximum duration of 180 simulated seconds.
The agent succeeds by remaining outside every pursuer's scanning sector for sixty consecutive simulated seconds within this limit.
Each sector has a 12\,m radius and a total angle of $90^\circ$.
Any exposure resets the undetected timer; three continuous seconds inside the union of scanning sectors causes capture.
Leaving all sectors resets the capture timer.
Pursuers and their visible minimap sectors use the same sensing geometry.

\paragraph{Aquatic.}
Completion depends on the water-task subtype.
Swimming goals use a 3\,m horizontal tolerance and a 0.8\,m depth tolerance; the submerged-entry task, for example, requires a depth of $6\pm0.8$\,m.
Boat-arrival tasks normally require successful boarding and arrival within 7\,m of the destination.
Swim-to-boat tasks instead end near the stern platform, while boarding tasks require the boarding transition itself.
Shore-arrival tasks require grounded walking on land, above the water surface, within 2.5\,m of the endpoint.
The beaching variant additionally checks that the boat is grounded and its full footprint is on land.
Shore-to-platform transport-choice variants accept surface arrival by swimming or boat.
The submarine variant requires crossing the pipe mouth and reaching its interior goal section with valid depth, alignment, and hull clearance.

\paragraph{Aerial.}
The agent must board the aircraft, take off, satisfy the configured aerial objective, and land at the designated recovery site.
Flight waypoints are ordered three-dimensional gates, with radii of 65\,m for planes and 16\,m for helicopters.
Approach objectives require coming within 60\,m of the moving aircraft; hover objectives require three continuous seconds within 12\,m of the target at a speed no greater than 3\,m/s.
Landing requires ground contact at the correct runway or helipad after a recorded takeoff and landing, speed at most 1.5\,m/s, throttle below 5\%, and two continuous seconds stopped.
Helicopters must also be within 12\,m of the designated landing point.
Crashing, exiting the task aircraft, or leaving the allowed flight region causes failure.
An aircraft-approach task also fails if the target leaves its task airspace before the approach condition is met.

\paragraph{Shooting and Sniping.}
The evaluator transforms the reticle's impact point back into the annotated image coordinate system, accounting for camera movement and scope magnification.
Ordinary shooting tests the annotated target box and any required target-part mask; sniping tests the corresponding target polygons.
In both cases, exclusion regions are removed from the accepted hit region, and head-specific tasks use the annotated head region.
Single-target shooting and sniping allow one shot: a hit succeeds and a miss fails.
In multi-target sniping, a task with $K$ targets supplies exactly $K$ bullets and requires $K$ distinct valid hits.
An unscoped shot, a miss, or a repeated hit on an already cleared target immediately fails a multi-target task.
Single-target sniping uses seeded dispersion for unscoped fire where supported, while scoped fire uses a point impact.

\paragraph{Interaction budgets and termination precedence.}
Online episodes allow up to 300 interaction rounds, and image-based tasks allow up to 20.
Exhausting the round budget without success yields zero credit.
Following requires the agent to maintain the required proximity until the NPC reaches its endpoint.
Task-specific failures, including capture, aircraft destruction, and the target reaching its endpoint in Interception, take precedence over completion.

\subsection{Agent Action Space}
\label{sec:appendix_actions}
The agent controls the environment through a shared keyboard--mouse interface.
Each response specifies one action object with a required \texttt{keys} array and an optional \texttt{mouse} object.
Keyboard entries pair a key with its hold duration; mouse entries specify movement, scrolling, or button state changes.
The conventional GUI agent emits this object as JSON text; the harness agent places it in \texttt{game\_step.action}, alongside a unique \texttt{request\_id}.
Table~\ref{tab:action_space} summarizes the interface.

\begin{table}[t]
    \centering
    \small
    \setlength{\tabcolsep}{3pt}
    \begin{tabular}{@{}p{0.36\columnwidth}p{0.59\columnwidth}@{}}
        \hline
        Field & Parameters \\
        \hline
        \texttt{keys} & List of \texttt{key} and \texttt{duration\_seconds} pairs \\
        \texttt{mouse.move} & Relative pixel displacement $(d_x,d_y)$ \\
        \texttt{mouse.scroll} & Direction (up/down), integer steps $1$--$6$ \\
        \texttt{mouse.buttons} & Left/right button with down/up event \\
        \texttt{wait\_seconds} & Timed wait with an empty keyboard array \\
        \hline
    \end{tabular}
    \caption{Model-facing action schema. Mouse movement and scrolling are mutually exclusive within a response; button events may accompany either.}
    \label{tab:action_space}
\end{table}

\paragraph{Keyboard inputs.}
The permitted keys are \texttt{w}, \texttt{a}, \texttt{s}, \texttt{d}, \texttt{Shift}, \texttt{Control}, \texttt{Space}, \texttt{e}, \texttt{f}, \texttt{q}, \texttt{c}, \texttt{v}, \texttt{g}, and \texttt{Numpad4/5/6/8}.
Uppercase single-letter inputs and \texttt{Ctrl} are normalized to their canonical names.
Each key appears at most once per response and has a finite duration $d\in(0,d_{\max}]$, with a default limit of $d_{\max}=1$ second.
For native 3D tasks, the duration is rounded upward to simulation frames at 60\,Hz, and all specified keys start together or continue an existing hold.
Keys are released independently as their durations expire; the turn duration is the maximum scheduled hold duration rather than their sum.
There is no within-turn sequence of delayed key presses.

\paragraph{Mouse inputs.}
All observations use a $1280\times720$ viewport.
Mouse movement uses relative pixel displacement, with $-640\leq d_x\leq640$ and $-360\leq d_y\leq360$; finite fractional values are permitted.
Positive $d_x$ looks right and positive $d_y$ looks up.
Repeating a nonzero displacement changes the current view again.
The aiming center in shooting and sniping remains at $(640,360)$; native 3D tasks apply camera sensitivity rather than guaranteeing exact centering of an image point.
The agent can also scroll up or down by one to six steps, or issue down/up events for the left and right buttons, with at most one event per button in a turn.
Mouse inputs have no duration parameter and do not advance simulated time.

\paragraph{Persistent input state.}
In native 3D tasks, keys held through the end of a turn remain held while inference is paused.
Including the same key in the next response continues the hold; omitting it releases it before simulation resumes.
Mouse buttons remain in their current state until an explicit release or episode cleanup.
View orientation persists across turns, and no cursor coordinate is supplied as additional observation text.
An empty \texttt{keys} array releases previous keyboard holds.
With no mouse operation or wait, this produces a zero-time no-op that still consumes an interaction step.
The optional \texttt{wait\_seconds} field accepts a finite duration in $(0,1]$ seconds with \texttt{keys:[]}.
Instantaneous mouse operations occur first, then the simulation advances for the requested wait; image-based tasks consume a turn without advancing scene time.
All held inputs are released when an episode ends.

\Needspace{13\baselineskip}
\paragraph{Example.}
The following native action first applies a relative mouse movement of 120 pixels right and 30 pixels up, then holds \texttt{w} for one second and \texttt{Space} for 0.1 seconds concurrently.
The scheduled turn duration is one second, and the next screenshot is returned after execution or earlier task termination.
\begin{center}
\begin{BVerbatim}[fontsize=\footnotesize]
{
  "keys": [
    {"key": "w", "duration_seconds": 1},
    {"key": "Space", "duration_seconds": 0.1}
  ],
  "mouse": {"move": {"dx": 120, "dy": 30}}
}
\end{BVerbatim}
\end{center}

\paragraph{Image-based aiming tasks.}
Shooting and sniping use the mouse portion of the same schema; agents are instructed to submit an empty keyboard array, and well-formed keyboard inputs are ignored by the executor.
Ordinary shooting supports aiming movement and left-button down to fire.
Sniping additionally supports right-button down/up for scope control, left-button up, and scrolling to adjust magnification.
Each scroll step maps to 160\,ms of the aiming engine's zoom operation, without advancing a simulation clock.
A shot terminates a single-target task; multi-target sniping continues until all targets are cleared or a failure occurs, as specified in Appendix~\ref{sec:appendix_outcomes}.
Each response consumes one of the twenty available interaction steps.
The prompt requests one aiming operation per turn; the current executor also accepts composite mouse inputs without an intermediate observation.

\paragraph{Validation.}
Each action object is validated before execution.
Unknown fields, duplicate keys or buttons, invalid durations, out-of-range mouse displacements, simultaneous movement and scrolling, and waiting with a nonempty keyboard array are rejected.
For conventional GUI responses, the parser also accepts a single JSON code block enclosing the entire response, without surrounding explanation, although the prompt requests plain JSON.

% Freeze the prompt and executor together for the reported release. The current
% image-based prompt requests atomic operations, while stepStatic enables
% composite execution; do not describe atomicity as an enforced restriction.

%\subsection{Timing and Token Statistics}
%\label{sec:appendix_timing_statistics}
%Table~\ref{tab:evaluation_efficiency} supplements the evaluation-setting comparisons in Section~\ref{sec:evaluation_settings}.
%It reports mean interaction rounds, simulated game time, requested action duration, and input/output tokens on the same 103 online tasks.
%Unpaused game time includes simulation advance during inference; it is distinct from the sum of scheduled key-hold durations.
%The half-second setting permits 600 rounds, while the one- and two-second settings permit 300.

\begin{table}[t]
    \centering
    \small
    \setlength{\tabcolsep}{2.6pt}
    \begin{tabular}{@{}ccrrrrr@{}}
        \toprule
        Cap & Pause & Rounds & Game (s) & $\bar\delta$ (s) & In (M) & Out (k) \\
        \midrule
        0.5 & Yes & 121.2 & 59.7 & 0.494 & 11.20 & 7.66 \\
        1 & Yes & 62.6 & 56.9 & 0.916 & 4.32 & 4.27 \\
        2 & Yes & 42.5 & 61.5 & 1.468 & 2.50 & 3.10 \\
        1 & No & 31.8 & 448.3 & 0.914 & 1.83 & 2.10 \\
        \bottomrule
    \end{tabular}
    \caption{Timing and token usage on the same 103 tasks. Rounds, game time, and input/output tokens are task means. $\bar\delta$ averages the maximum requested key/wait duration within each executed action. Cap is in seconds; M/k denote millions/thousands of tokens. Unpaused game time includes simulation advance during inference.}
    \label{tab:evaluation_efficiency}
\end{table}

\subsection{Model Configurations}
\label{sec:appendix_model_configurations}

Table~\ref{tab:model-configurations} lists the reasoning effort settings
used for the results in Table~\ref{tab:gtabench-main} and
Figures~\ref{fig:performance_radar} and~\ref{fig:performance_tokens}.
Under both the JSON and Codex evaluation protocols, all models use a
1M-token context window, a maximum of 32,768 output tokens per response,
and a sampling temperature of 0.7. Each configuration was evaluated using three independent random seeds.

\begin{table}[htbp]
    \centering
    \begin{tabular}{ll}
        \hline
        \textbf{Model} & \textbf{Reasoning effort} \\
        \hline
        GPT-6-Astra              & \texttt{high} \\
        GPT-6.1-Sol              & \texttt{high} \\
        DeepSeek-V4.1-Flash-1017 & \texttt{max} \\
        GLM-5.3-Flash            & \texttt{max} \\
        Kimi-K3                  & \texttt{max} \\
        Qwen3.8-27B              & \texttt{xhigh} \\
        \hline
    \end{tabular}
    \caption{Reasoning effort by model under the JSON and Codex
    evaluation protocols.}
    \label{tab:model-configurations}
\end{table}

\begin{figure*}[tb]
    \centering
    \includegraphics[width=\textwidth]{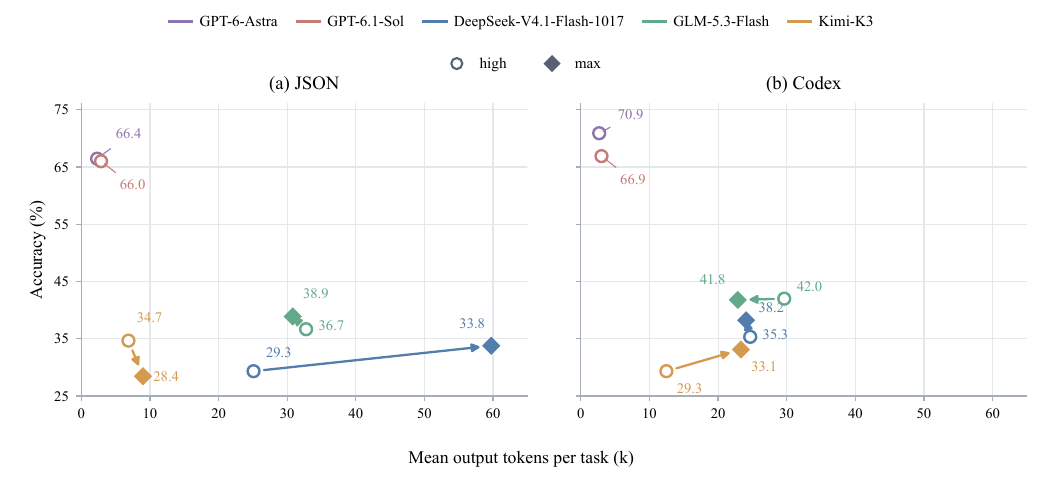}
    \caption{Accuracy and output-token usage across reasoning-effort settings on 150 tasks. GPT models serve as \texttt{high}-effort references.}
    \label{fig:reasoning_effort}
\end{figure*}

\paragraph{Reasoning-effort comparison.}
Figure~\ref{fig:reasoning_effort} compares requested \texttt{high} and \texttt{max} effort for DeepSeek, GLM, and Kimi under both protocols, using the same 150 task IDs and the token-accounting rules above.
For these three models, each \texttt{high} result comes from one completed run; \texttt{max} retains the main-table aggregation.
GPT-6-Astra and GPT-6.1-Sol are also shown at \texttt{high} as reference points using the main-table configurations and Figure~\ref{fig:performance_tokens} token measurements.
These historical batches are not a controlled paired ablation. Higher requested effort does not consistently improve accuracy or reduce output-token usage.

\subsection{Human Evaluation}
\label{sec:appendix_human_evaluation}
We recruited three experienced players who had previously completed \textit{Grand Theft Auto V}.
Participants played directly through the task interfaces using standard gameplay controls.
Each participant attempted all 150 benchmark tasks exactly once, with no prior exposure to those tasks.
Table~\ref{tab:gtabench-main} reports the mean accuracy and population standard deviation (in percentage points) across the three participants.
Overall accuracy weights all 150 tasks equally.

\section{RL Environment Parameterization}
\label{sec:appendix_rl_parameters}
A template fixes the scene, available assets, task logic, outcome rules, and episode budget.
For image-based shooting, it fixes an annotated image pool and the aiming interface.
The parameters in Table~\ref{tab:rl_parameters} are numeric quantities or categorical choices whose sampling domains can be restricted by the user.
Table~\ref{tab:rl_family_mapping} marks the parameters exposed by each template.
Shared quantities use the same symbols across templates.
All domains are intersected with the template's valid locations, assets, and controller constraints.
Positions use $(x,y,z)$, with $(x,z)$ spanning the horizontal plane and $y$ giving vertical elevation; underwater depth is measured below the local water surface.

\begin{table*}[t]
    \centering
    \small
    \setlength{\tabcolsep}{4pt}
    \renewcommand{\arraystretch}{1.12}
    \begin{tabular}{@{}p{0.18\textwidth}p{0.13\textwidth}p{0.64\textwidth}@{}}
        \toprule
        Parameter & Type & User-controlled sampling domain \\
        \midrule
        Start $p_s$ & Position & Coordinates sampled within $[x_s^-,x_s^+]\times[y_s^-,y_s^+]\times[z_s^-,z_s^+]$, intersected with valid spawn locations in the fixed scene. \\
        Destination $p_g$ & Position & Coordinates sampled within $[x_g^-,x_g^+]\times[y_g^-,y_g^+]\times[z_g^-,z_g^+]$, intersected with reachable goal locations. The $y$ interval constrains elevation and thus underwater depth. For following and interception, this is the target's destination. \\
        Heading $\psi$ & Real angle & An interval $[\psi_{\min},\psi_{\max}]$ in degrees for the agent's initial yaw. Other pose components follow template initialization rules. \\
        Route length $\ell$ & Real distance & An interval $[\ell_{\min},\ell_{\max}]$ in meters for the path length. Applies to the agent's reference route for navigation and the target's route for following or interception. \\
        Turn count $k$ & Integer & Integers in $[k_{\min},k_{\max}]$ under the template's turn-count convention, applied to the same route as $\ell$. \\
        Transport $m$ & Category & An explicit allowed set $\mathcal{M}\subseteq\{\text{land},\text{water},\text{air}\}$ specifies the transport domain, filtered by template support. Movement $u$ specifies the concrete mode within that domain. \\
        Movement $u$ & Category & An allowed set $\mathcal{U}\subseteq\{\text{walk},\text{drive},\text{swim},\text{dive},\text{boat},\text{plane},\text{helicopter}\}$. With transport fixed, the \texttt{movement} field restricts the concrete movement mode to choices compatible with both that domain and the task template. \\
        Actor speed $v$ & Real speed & An interval $[v_{\min},v_{\max}]$ in meters per second for the target or pursuers; the same bounds apply to each pursuer. \\
        Initial distance $d$ & Real distance & An interval $[d_{\min},d_{\max}]$ in meters between the agent and each target or pursuer at initialization. Actor positions must also satisfy scene constraints. \\
        Pursuer count $n$ & Integer & Integers in $[n_{\min},n_{\max}]$, within the evasion template's supported capacity. \\
        Image--target $i$ & Category & An explicit index set $J_i\subseteq\{1,\ldots,N_T\}$, where $N_T$ is the number of annotated shooting pairs in the template manifest. For example, $J_i=\{1,3,5\}$ allows only those three pairs. \\
        Aim offset $\delta$ & Real vector & Horizontal and vertical pixel-offset intervals relative to the annotated target center. The resulting initial aim must remain inside the viewport. \\
        \bottomrule
    \end{tabular}
    \caption{Sampling parameters and their admissible input types. Users set intervals, spatial regions, or candidate sets; the generator samples valid instances within those restrictions.}
    \label{tab:rl_parameters}
\end{table*}

\begin{table*}[!t]
    \centering
    \small
    \setlength{\tabcolsep}{3pt}
    \renewcommand{\arraystretch}{1.12}
    \resizebox{\textwidth}{!}{%
    \begin{tabular}{@{}l*{10}{c}@{}}
        \toprule
         & \multicolumn{6}{c}{Navigation} & \multicolumn{3}{c}{Interaction} & Gun Use \\
        \cmidrule(lr){2-7}\cmidrule(lr){8-10}\cmidrule(lr){11-11}
        Sampling parameter & \shortstack{Road} & \shortstack{On-Foot} & \shortstack{Aquatic} & \shortstack{Vertical} & \shortstack{Aerial} & \shortstack{Indoor} & \shortstack{Following} & \shortstack{Interception} & \shortstack{Evasion} & \shortstack{Shooting} \\
        \midrule
        Start $p_s$ & $\checkmark$ & $\checkmark$ & $\checkmark$ & $\checkmark$ & $\checkmark$ & $\checkmark$ & $\checkmark$ & $\checkmark$ & $\checkmark$ & --- \\
        Destination $p_g$ & $\checkmark$ & $\checkmark$ & $\checkmark$ & $\checkmark$ & $\checkmark$ & $\checkmark$ & $\checkmark$ & $\checkmark$ & --- & --- \\
        Heading $\psi$ & $\checkmark$ & $\checkmark$ & $\checkmark$ & $\checkmark$ & $\checkmark$ & $\checkmark$ & $\checkmark$ & $\checkmark$ & $\checkmark$ & --- \\
        Route length $\ell$ & $\checkmark$ & $\checkmark$ & $\checkmark$ & $\checkmark$ & $\checkmark$ & $\checkmark$ & $\checkmark$ & $\checkmark$ & --- & --- \\
        Turn count $k$ & $\checkmark$ & $\checkmark$ & $\checkmark$ & $\checkmark$ & $\checkmark$ & $\checkmark$ & $\checkmark$ & $\checkmark$ & --- & --- \\
        Transport $m$ & --- & --- & --- & --- & --- & --- & $\checkmark$ & --- & $\checkmark$ & --- \\
        Movement $u$ & --- & --- & $\checkmark$ & --- & $\checkmark$ & --- & $\checkmark$ & --- & $\checkmark$ & --- \\
        Actor speed $v$ & --- & --- & --- & --- & --- & --- & $\checkmark$ & $\checkmark$ & $\checkmark$ & --- \\
        Initial distance $d$ & --- & --- & --- & --- & --- & --- & $\checkmark$ & $\checkmark$ & $\checkmark$ & --- \\
        Pursuer count $n$ & --- & --- & --- & --- & --- & --- & --- & --- & $\checkmark$ & --- \\
        Image--target $i$ & --- & --- & --- & --- & --- & --- & --- & --- & --- & $\checkmark$ \\
        Aim offset $\delta$ & --- & --- & --- & --- & --- & --- & --- & --- & --- & $\checkmark$ \\
        \bottomrule
    \end{tabular}%
    }
    \caption{Range-controlled sampling parameters for each task template. A checkmark indicates a parameter sampled within user-specified bounds or candidate sets; a dash indicates that it is fixed by the template or inapplicable. Transport selects land, water, or air; movement selects a supported concrete mode within that domain. For following and interception, $p_g$, $\ell$, and $k$ describe the target route. Parameter types and feasible domains are defined in Table~\ref{tab:rl_parameters}.}
    \label{tab:rl_family_mapping}
\end{table*}

\paragraph{Jointly feasible sampling.}
Numeric bounds and option sets constrain a distribution of samples within a fixed template.
For a template supporting land and water transport, $\mathcal{M}=\{\text{land},\text{water}\}$ samples between these domains, while $\mathcal{M}=\{\text{land}\}$ fixes land transport.
Within a selected domain, $\mathcal{U}$ controls the concrete movement mode: land supports walking or driving, water supports swimming, diving, or boats, and air supports planes or helicopters, subject to template restrictions.
For example, \texttt{transport: [water]} with \texttt{movement: [boat]} fixes boat navigation; \texttt{movement: [swim, boat]} allows either supported mode.
If \texttt{movement} is omitted, the generator uses the template's movement modes compatible with the transport restriction.
Endpoint choices, path length, and turn count must admit a compatible route, and transport options must support the selected locations.
The generator searches for valid routes and placements within these restrictions, rejects incompatible proposals, and reports failure if its generation budget is exhausted.
Equal numeric bounds, a fixed location, or a singleton option set hold a parameter constant.

\paragraph{Navigation and moving actors.}
Navigation templates use $p_s$, $p_g$, $\psi$, $\ell$, and $k$ to generate the initial pose and reference route.
Their transport domain is fixed: Road uses land, Aquatic uses water, and Aerial uses air.
Road uses driving, while On-Foot, Vertical, and Indoor retain walking as their template-defined movement.
Aquatic and Aerial expose $u$ to select a concrete mode within their fixed water or air domain.
Following and Evasion expose both $m$ and $u$ within their supported choices; Interception fixes land transport and driving.
Elevation constraints on $p_g$ and the template's traversable geometry determine valid vertical, aquatic, and aerial routes.
Following and Interception use the same start and heading parameters for the agent, but $p_g$, $\ell$, and $k$ describe the target route.
The target starts at a valid location satisfying $d$ and moves at the sampled speed $v$.
Evasion instead samples $n$ pursuers, with each pursuer's speed and initial distance drawn within the specified $v$ and $d$ bounds.
It imposes no destination or reference route on the agent.
Pursuer identities, placement rules, sensing configuration, and behavior controllers are supplied by the template.
Following-distance tolerances, interception conditions, and escape timers remain fixed template rules.

\paragraph{Image-based sampling.}
The categorical choice $i$ selects an existing image--target pair, preserving the correspondence between visible content and hit annotations.
The generator then samples $\delta$ subject to the selected pair's valid image bounds.
Different pair choices and aiming offsets yield different shooting samples within the same template.
Sniping is evaluation-only and is excluded from environment generation.

\paragraph{From ranges to a concrete sample.}
As an illustrative evasion configuration, a user could restrict pursuer count to $\{1,2,3\}$, initial distance to $[20,40]$ meters, and speed to $[2,4]$ meters per second.
A generation seed $\xi$ controls the choices made within these bounds.
The generator saves the sampled values, resolved route and actor placements or aiming state, and template and environment versions.
These resolved values identify the concrete instance; the input ranges describe the family of instances from which it was sampled.

\subsection{On-Foot RL Training Setup}
\label{sec:appendix_lora_rl}

\paragraph{Generated training split.}
We use generation seed 20261027 to construct 50 on-foot instances.
Route lengths and turn counts follow the empirical joint distribution of the 16 benchmark on-foot tasks, with a 5\% length tolerance for new routes.
The resulting routes span 88.9--131.4 meters and zero to four turns; 19 instances include destination landmarks such as motorcycles, sedans, vans, or convertibles.
The generator checks route feasibility using the movement controller and enforces spatial separation from the benchmark on-foot routes.
Training and evaluation use distinct task instances within the same city and controllers.

\paragraph{Optimization and interaction.}
We train rank-32 LoRA adapters with $\alpha=64$ on the language model's attention query and value projections; the vision encoder and base model weights remain frozen.
GRPO uses AdamW with a constant learning rate of $10^{-6}$ and clipping ratio 0.2.
Each update uses four tasks and eight rollouts per task, collecting 32 concurrent trajectories; optimization uses eight H200 GPUs.
Training uses the native Codex history and automatic compaction, the same game rules.
Instructions appear in the initial screenshot, and inference pauses the environment.
The context window is 1M tokens, generation is capped at 32,768 tokens per response, actions at one second, and episodes at 300 interaction rounds.

\subsection{Reward Implementation Details}
\label{sec:appendix_rl_reward}
We detail the terminal reward defined in Eq.~\ref{eq:rl_terminal_reward} for the on-foot GRPO experiment.

\paragraph{Normalized distance progress.}
Let $P(\tau)$ denote the clipped fractional distance reduction in Eq.~\ref{eq:rl_terminal_reward}.
Normalization makes equal fractional reductions comparable across route lengths.
Moving farther from the destination yields zero progress, and revisiting locations provides no accumulated bonus: only the final distance is used.
The factor $(1-S)$ applies the distance bonus only to unsuccessful rollouts.
Thus a successful rollout receives either $1$ or $0.9$, while an unsuccessful rollout receives at most $0.3$.

\paragraph{Traversable-distance measurement.}
Distances are approximated on a one-meter, eight-neighbor walking grid using A* search with the environment's collision and surface-height checks.
The search is independent of the displayed purple route and permits physically feasible shortcuts.
An unreachable or unsupported-height final position receives $P=0$.
Distance computation failures are treated as invalid reward evidence, not as zero task progress.
These measurements are host-side training signals and are not included in the agent's observations.

\paragraph{Round penalty and terminal scoring.}
The $0.1$ deduction is applied once when $N>100$; it is not charged per additional round and is not applied at exactly 100 rounds.
$N$ includes executed-action rounds and feedback rounds that resume an unfinished agent after a premature final response.
The reward is assigned once at the end of the rollout, with no intermediate distance rewards.
An unsuccessful rollout ending at a game-time or interaction limit retains its final-distance progress term.
Benchmark evaluation continues to report binary task success alone.

\subsection{Native GTA V Evaluation}
\label{sec:appendix_native_gta_transfer}
The native GTA V comparison uses 100 tasks across the nine 3D navigation and interaction categories, excluding shooting and sniping.
Table~\ref{tab:native_gta_mission_sources} maps each category to its source missions, grouping alternative heist approaches under the same mission name.
% Generated by scripts/generate_native_gta_mission_table.py.
\begin{table*}[t]
    \centering
    \begingroup
    \footnotesize
    \setlength{\tabcolsep}{4pt}
    \renewcommand{\arraystretch}{1.12}
    \begin{tabular}{@{}>{\raggedright\arraybackslash}p{0.17\textwidth}
                        >{\centering\arraybackslash}p{0.06\textwidth}
                        >{\raggedright\arraybackslash}p{0.72\textwidth}@{}}
        \toprule
        \textbf{Task category} & \textbf{Tasks} & \textbf{Source missions in GTA V} \\
        \midrule
        On-Foot Navigation & 9 & Architect's Plans; Bury the Hatchet; Chop; Eye in the Sky; Franklin and Lamar; Friend Request; Mr. Richards; The Construction Assassination; The Long Stretch. \\
        \addlinespace[3pt]
        Road Driving & 21 & Architect's Plans; By the Book; Casing the Jewel Store; Chop; Cleaning Out the Bureau; Deep Inside; Father/Son; Friends Reunited; Meltdown; Mr. Philips; Reuniting the Family; Scouting the Port; The Big Score; The Bureau Raid; The Bus Assassination; The Merryweather Heist. \\
        \addlinespace[3pt]
        Aquatic Navigation & 12 & Daddy's Little Girl; Derailed; Hood Safari; Minisub; The Merryweather Heist. \\
        \addlinespace[3pt]
        Vertical Navigation & 7 & Casing the Jewel Store; Complications; Minisub; The Bureau Raid; The Time's Come. \\
        \addlinespace[3pt]
        Aerial Navigation & 9 & Minor Turbulence; Monkey Business; Nervous Ron; Predator; The Bureau Raid; The Merryweather Heist; Three's Company. \\
        \addlinespace[3pt]
        Indoor Navigation & 9 & Crystal Maze; Friend Request; Meltdown; The Big Score; The Bureau Raid; The Paleto Score. \\
        \addlinespace[3pt]
        Following & 7 & Cleaning Out the Bureau; Eye in the Sky; Franklin and Lamar; Minor Turbulence; Repossession; Surveying the Score. \\
        \addlinespace[3pt]
        Interception & 14 & Caida Libre; Chop; Derailed; Doting Dad; Fame or Shame; Father/Son; Legal Trouble; Marriage Counseling; Predator; Repossession; Something Sensible; The Time's Come. \\
        \addlinespace[3pt]
        Evasion & 12 & Cargobob; Dead Man Walking; Franklin and Lamar; Hood Safari; Legal Trouble; Pack Man; The Big Score; The Good Husband; The Long Stretch; The Wrap Up. \\
        \midrule
        \textbf{Total} & \textbf{100} & 49 unique source missions. \\
        \bottomrule
    \end{tabular}
    \caption{Task counts and source missions for the nine categories in the 100-task native GTA V evaluation. A mission may supply multiple tasks or categories; alternative heist approaches are grouped under the same mission name.}
    \label{tab:native_gta_mission_sources}
    \endgroup
\end{table*}

\paragraph{Settings.}
The model uses the same screenshot--keyboard--mouse interface and turn-based evaluation procedure as \benchmark, with the Codex harness used in \benchmarkagi.
The game pauses during model inference; the requested timed inputs are then executed, and the resulting screenshot is returned for the next decision.

\paragraph{Human verification.}
Task completion is manually verified by reviewing the model's gameplay recording from the handoff onward.
Reviewers assign success when the video shows that the model reaches the game's native task checkpoint or triggers an explicit task-completion cue.
For navigation, one such cue is the minimap changing from a displayed GPS route to no route after the player arrives at the marked destination.
Other evidence includes progression to the next native objective or an explicit mission-completion message.

\section{Error Annotation and Recovery Analysis}
\label{sec:appendix_error_annotation}
This appendix documents the first-error analysis in Section~\ref{sec:failure_modes}.
Table~\ref{tab:dynamic_error_cases} additionally illustrates the dynamic-control failures discussed in Section~\ref{sec:category_control}.
The annotation cohort contains one recorded \benchmark JSON-protocol run for each of six models on the same 150 task identities: 900 trajectories in total.
Our analysis includes 565 trajectories with a localized first error and a capability label.
For each, we identify the first task-relevant error, assign a capability category, and assess whether subsequent actions restore task progress.

\subsection{Capability Definitions}
\label{sec:appendix_error_taxonomy}
Table~\ref{tab:error_category_definitions} summarizes the common observable conditions for each of the four categories.
Labels describe failure mechanisms supported by observations and executed actions.
The four categories are Perception, Planning, Calibration, and Regulation. Each trajectory contributes at most one first error to the category distribution.
The 565 categorized first errors comprise 112 Perception, 96 Planning, 157 Calibration, and 200 Regulation cases.
Calibration and Regulation together account for 63.2\% of these errors. Table~\ref{tab:first_error_cases} illustrates evidence-based cases.

\begin{table}[H]
    \centering
    \footnotesize
    \setlength{\tabcolsep}{3pt}
    \renewcommand{\arraystretch}{1.12}
    \begin{tabular}{@{}>{\raggedright\arraybackslash}p{0.23\columnwidth}
                        >{\raggedright\arraybackslash}p{0.73\columnwidth}@{}}
        \toprule
        \textbf{Category} & \textbf{Common error conditions} \\
        \midrule
        \textbf{Perception}
        & Misreading visible objects, route geometry, or road connectivity; confusing targets or minimap markers; misaligning camera, map, and movement reference frames. \\
        \addlinespace[3pt]
        \textbf{Planning}
        & Skipping a required interaction or movement mode; choosing incompatible subgoals or an incorrect stage order; repeating an ineffective strategy despite sustained evidence of no progress. \\
        \addlinespace[3pt]
        \textbf{Calibration}
        & Selecting the wrong control axis or polarity; overshooting or undershooting because an input's magnitude or duration is inappropriate for the immediate adjustment. \\
        \addlinespace[3pt]
        \textbf{Regulation}
        & Initiating or ending a maneuver too early or late; failing to brake or stabilize during approach; neglecting changes in ego-motion, target distance, or threats. \\
        \bottomrule
    \end{tabular}
    \caption{Common annotation criteria for the four first-error categories. Calibration concerns the direction and size of an immediate input; Regulation concerns adjustment as the situation evolves. A collision, miss, or unsuccessful interaction alone does not establish its underlying mechanism.}
    \label{tab:error_category_definitions}
\end{table}

% Original observations and state evidence: figures/dynamic-error-cases/manifest.json.
\begin{table*}[p]
\centering
\footnotesize
\setlength{\tabcolsep}{3pt}
\renewcommand{\arraystretch}{1.08}
\begingroup
\newcommand{\dynamiccaseimage}[2]{\begin{minipage}{\linewidth}
    #1\par
    \vspace{2pt}\scriptsize #2\end{minipage}}
\begin{tabular}{@{}>{\raggedright\arraybackslash}m{0.11\textwidth}
                    >{\raggedright\arraybackslash}m{0.29\textwidth}
                    >{\raggedright\arraybackslash}m{0.29\textwidth}
                    >{\raggedright\arraybackslash}m{0.25\textwidth}@{}}
\toprule
\textbf{Category} & \textbf{Observation and action} & \textbf{Next observation} & \textbf{Failure mechanism} \\
\midrule
\textbf{Road navigation}
& \dynamiccaseimage{\includegraphics[width=\linewidth]{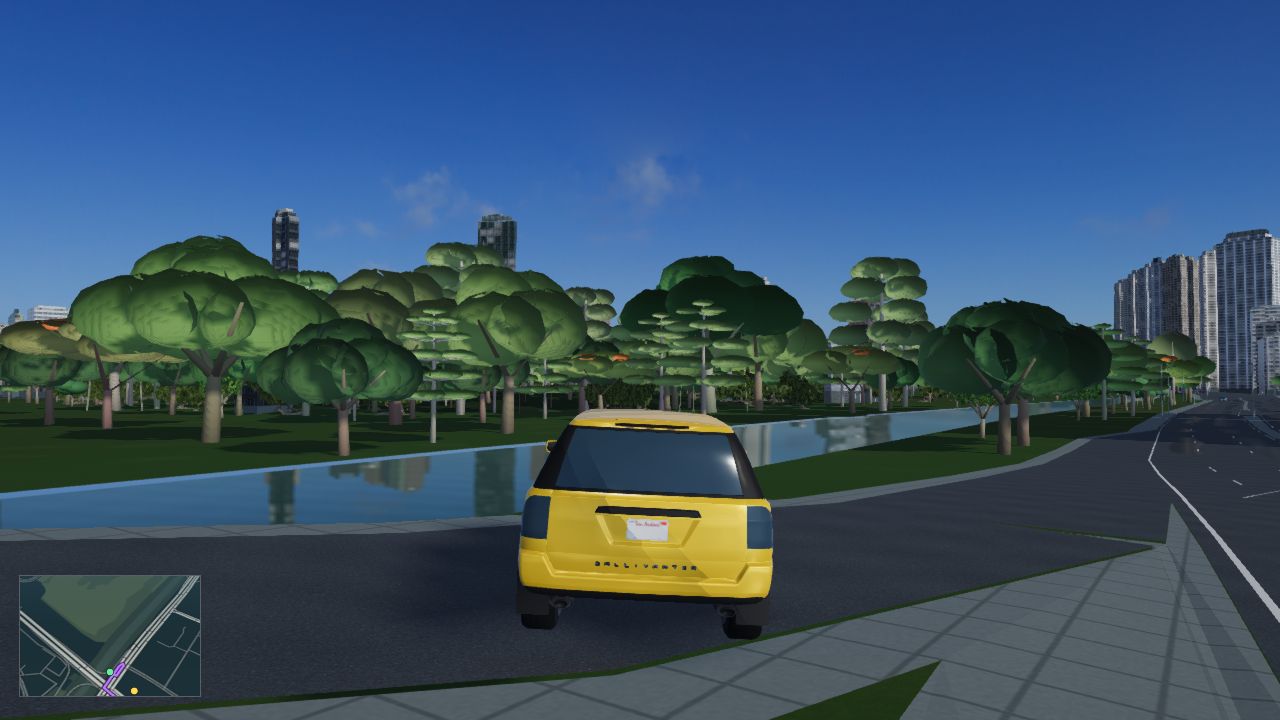}}{\textbf{Action:} W 1\,s + D 1\,s. Kimi-K3 approaches the bend at 24.2\,m/s.}
& \dynamiccaseimage{\includegraphics[width=\linewidth]{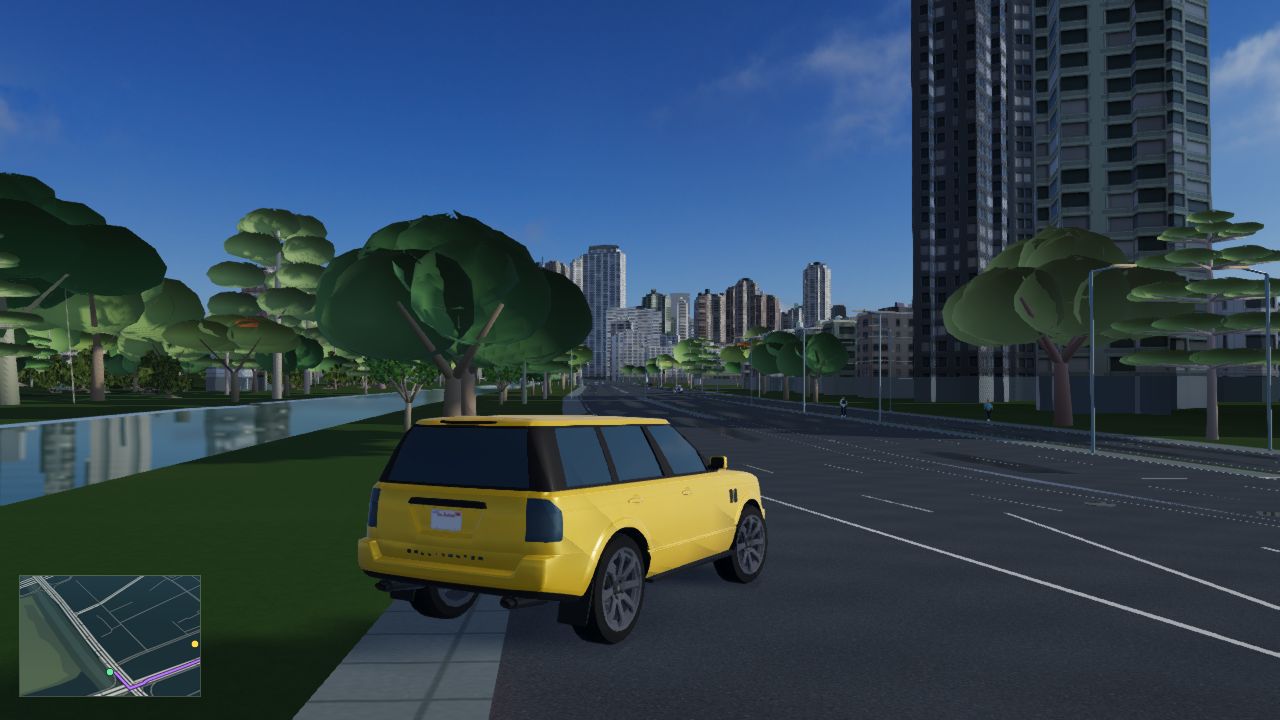}}{Speed rises to 30.5\,m/s as the vehicle turns approximately $67^\circ$, ending on the road edge and pavement.}
& \textbf{Excessive steering while accelerating.} A full-second steering input carries the vehicle onto the curb rather than aligning it with the road. The next action continues accelerating without a steering correction; the vehicle subsequently travels onto the grass. \\
\addlinespace[6pt]
\textbf{Interception}
& \dynamiccaseimage{\includegraphics[width=\linewidth]{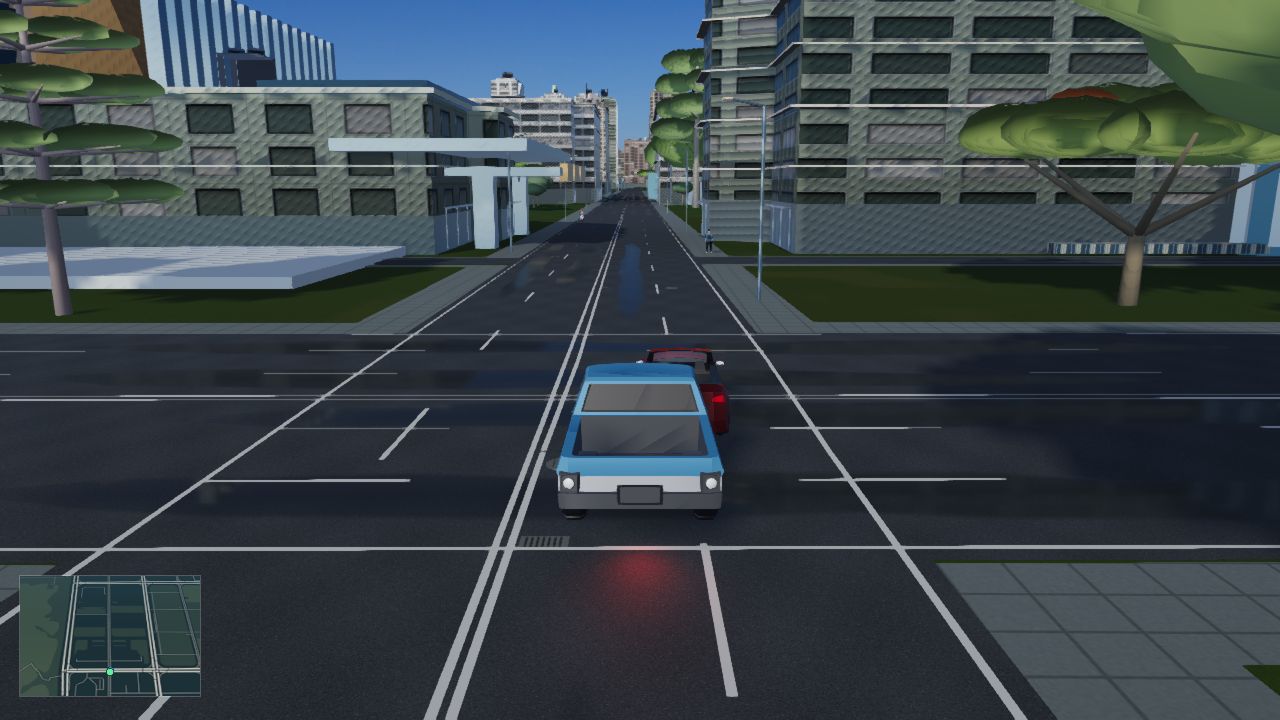}}{\textbf{Action:} W 0.8\,s + A 0.22\,s. Impact has slowed the target to 0.2\,m/s, but the player is still moving at 37.5\,m/s.}
& \dynamiccaseimage{\includegraphics[width=\linewidth]{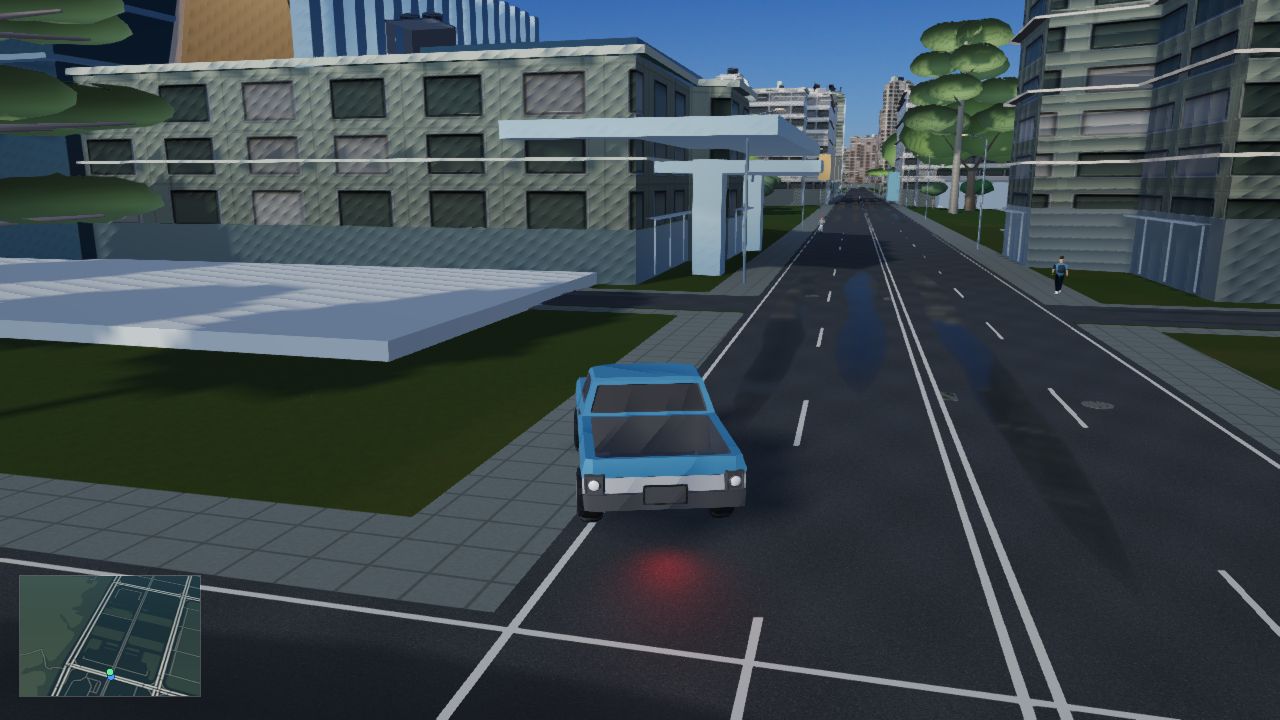}}{The player moves 20.4\,m and loses contact, still at 36.8\,m/s. No blocking hold accumulates.}
& \textbf{Excess speed after impact.} Rather than braking immediately to establish a blocking position, the agent continues applying throttle and steers away from contact. It brakes only afterward; the target later resumes motion and reaches its endpoint. \\
\addlinespace[6pt]
\textbf{Aerial navigation}
& \dynamiccaseimage{\includegraphics[width=\linewidth]{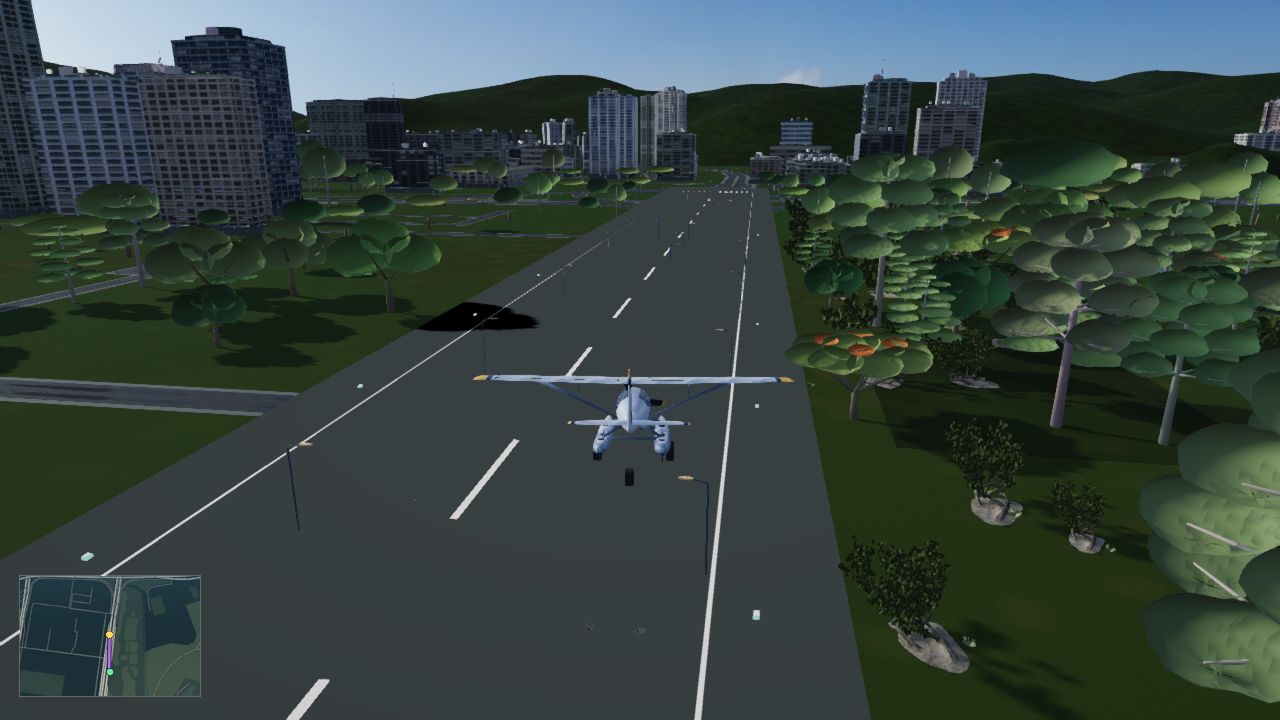}}{\textbf{Action:} S 1\,s. The aircraft is near the runway at approximately 18.6\,m height, descending at 9.3\,m/s.}
& \dynamiccaseimage{\includegraphics[width=\linewidth]{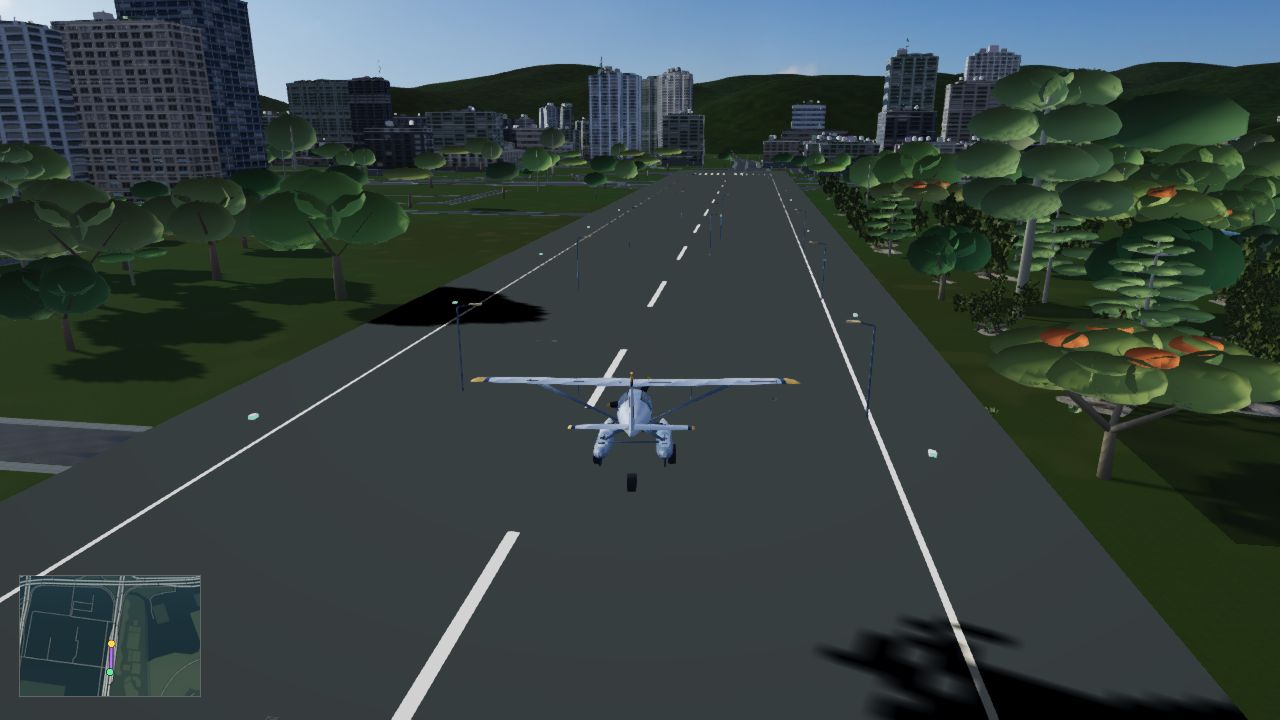}}{Height falls to approximately 9.4\,m while descent remains 9.2\,m/s.}
& \textbf{Late descent-rate correction.} Reducing throttle does not sufficiently arrest the rapid descent. The next action adds a brief pitch-up input, but touchdown still ends in a crash. \\
\addlinespace[6pt]
\textbf{Evasion}
& \dynamiccaseimage{\includegraphics[width=\linewidth]{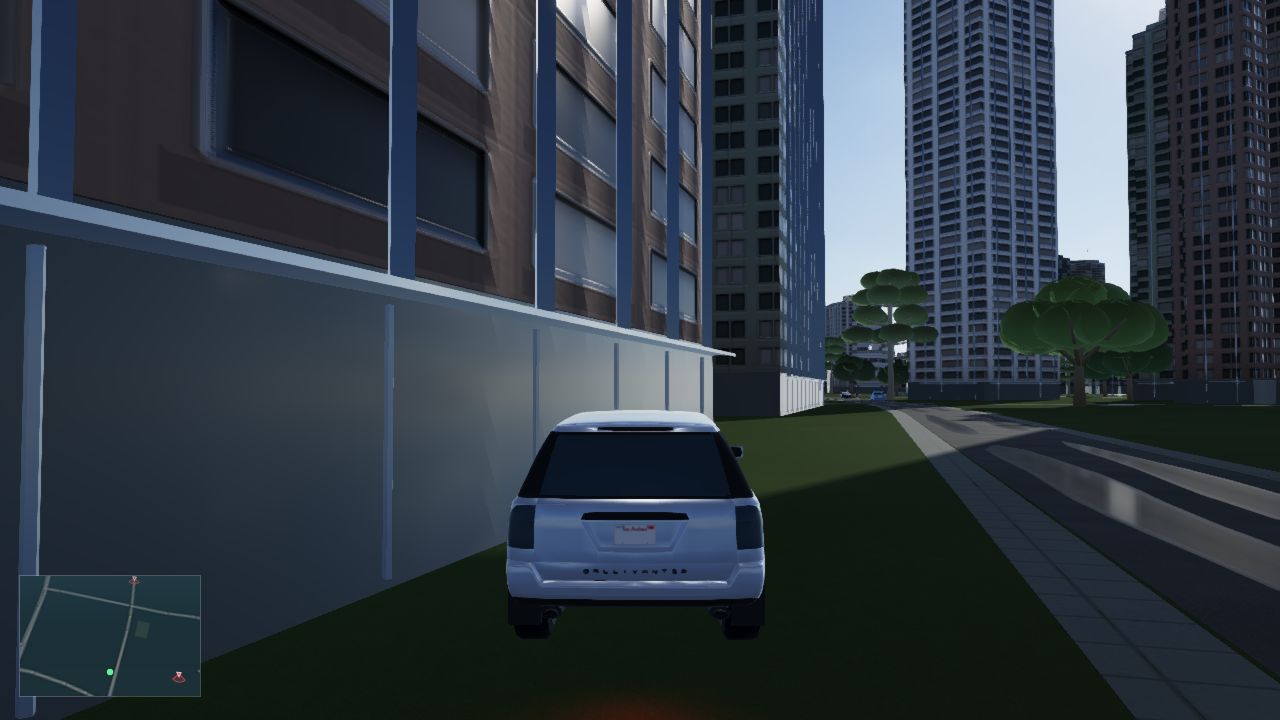}}{\textbf{Action:} W 1\,s + D 0.4\,s. The previous forward input ended in a building collision, reducing speed from 43.7\,m/s to zero.}
& \dynamiccaseimage{\includegraphics[width=\linewidth]{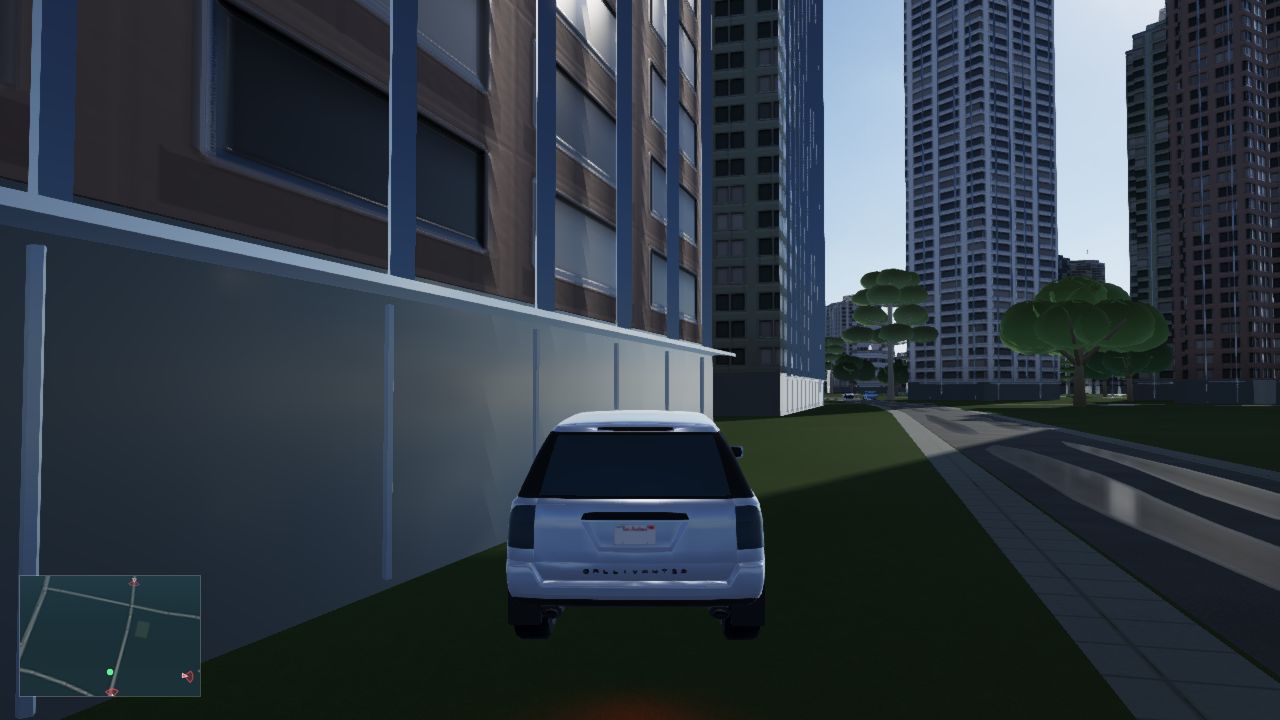}}{The vehicle remains pressed against the wall: the entire action produces zero displacement.}
& \textbf{Collision and wall entrapment.} Continued forward input with steering cannot free the vehicle from the building. The agent starts reversing only on the next decision, after spending a full second unable to move away from the wall. \\
\bottomrule
\end{tabular}
\endgroup
\caption{Control-error examples supporting Section~\ref{sec:category_control}. Road navigation uses Kimi-K3 under the JSON protocol; the other examples use GPT-6.1-Sol with Codex. Each pair contains the original observations immediately before and after the displayed action; simultaneous keys retain their recorded durations. }
\label{tab:dynamic_error_cases}
\end{table*}

\paragraph{First-error incidence.}
Let $N^{\mathrm{err}}_{m,c}$ be the number of trajectories for model $m$ whose identified first error belongs to capability $c$.
Each cell in Figure~\ref{fig:failure_capability_heatmap} reports $100N^{\mathrm{err}}_{m,c}/150$: the denominator is all 150 evaluated tasks for that model, rather than only its categorized first-error cases.
Thus, a row need not sum to 100\%.
The 63.2\% reported above instead describes the composition of the 565 categorized first errors pooled across models.

% Original screenshot and action provenance: figures/first-error-cases/manifest.json.
% Regenerate assets with: python scripts/export_first_error_examples.py
\begin{table*}[p]
    \centering
    \footnotesize
    \setlength{\tabcolsep}{3pt}
    \renewcommand{\arraystretch}{1.05}
    \begingroup
    \newcommand{\caseimage}[2]{%
        \begin{minipage}{\linewidth}
            #1\par
            \vspace{2pt}
            \centering\scriptsize #2
        \end{minipage}%
    }
    \newcommand{\caseaction}[1]{%
        \begin{minipage}{\linewidth}
            #1
        \end{minipage}%
    }
    \begin{tabular}{@{}>{\raggedright\arraybackslash}m{0.11\textwidth}
                        >{\raggedright\arraybackslash}m{0.31\textwidth}
                        >{\raggedright\arraybackslash}m{0.26\textwidth}
                        >{\raggedright\arraybackslash}m{0.28\textwidth}@{}}
        \toprule
        \textbf{Capability} & \textbf{Observation} & \textbf{Model output action} & \textbf{Why this is an error} \\
        \midrule
        \textbf{Perception}
        & \caseimage{\includegraphics[width=\linewidth]{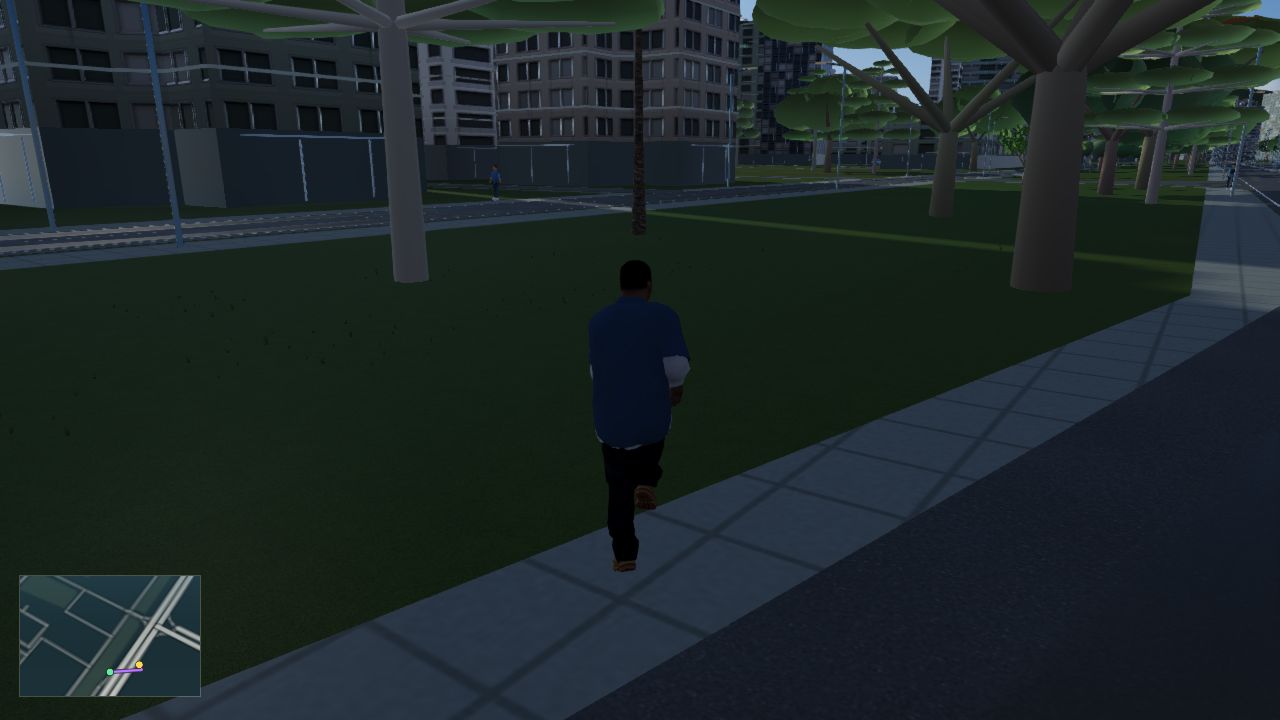}}{Qwen3.8-27B}
        & \caseaction{
\begin{lstlisting}[language={},basicstyle=\ttfamily\fontsize{7}{8}\selectfont,
                columns=fullflexible,keepspaces=true,breaklines=true,breakatwhitespace=false,
                showstringspaces=false,upquote=true,frame=none,
                aboveskip=0pt,belowskip=0pt]
{
  "keys": [
    {"key": "w",
     "duration_seconds": 1},
    {"key": "Shift",
     "duration_seconds": 1}
  ],
  "mouse": {"move": {"dx": 0, "dy": 0}}
}
\end{lstlisting}}
        & The model reads the minimap destination as ``very close, just ahead,'' although the purple route and yellow marker lie to the right of the player marker. It keeps running straight into the grass instead of correcting toward the indicated route. \\
        \addlinespace[4pt]
        \textbf{Planning}
        & \caseimage{\includegraphics[width=\linewidth]{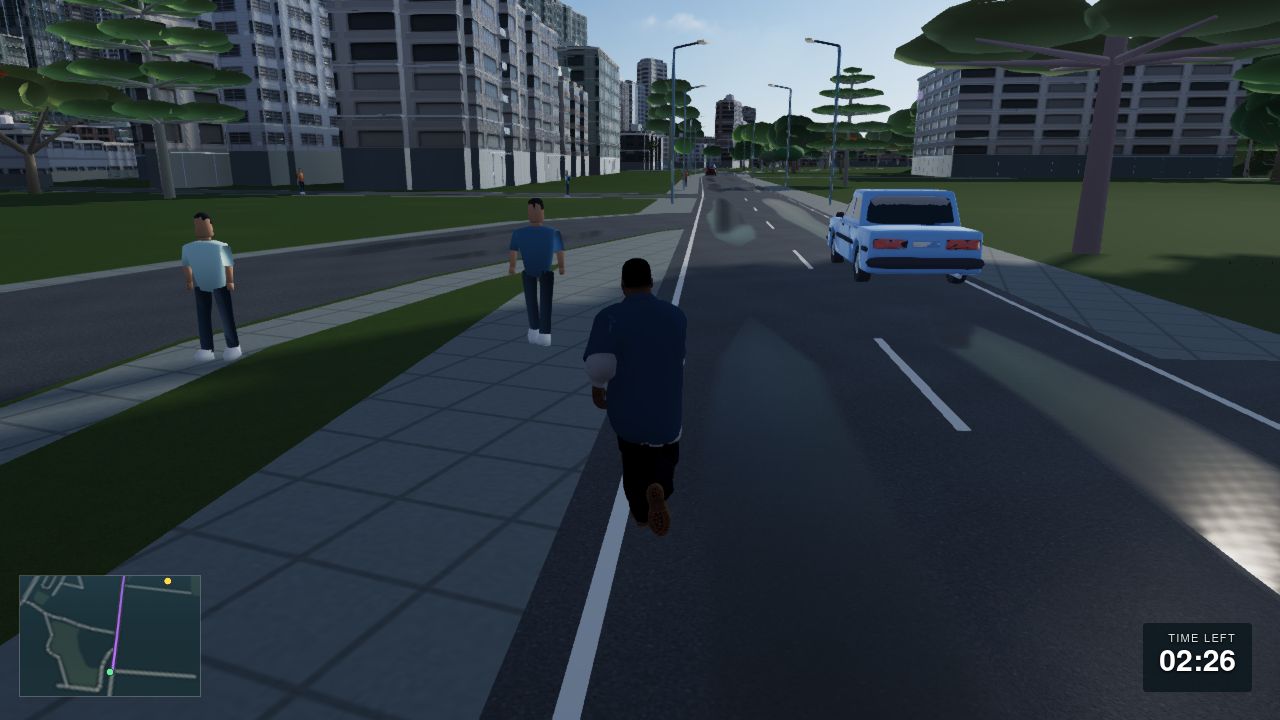}}{Qwen3.8-27B}
        & \caseaction{
\begin{lstlisting}[language={},basicstyle=\ttfamily\fontsize{7}{8}\selectfont,
                columns=fullflexible,keepspaces=true,breaklines=true,breakatwhitespace=false,
                showstringspaces=false,upquote=true,frame=none,
                aboveskip=0pt,belowskip=0pt]
{
  "keys": [
    {"key": "w",
     "duration_seconds": 1.0},
    {"key": "Shift",
     "duration_seconds": 1.0}
  ],
  "mouse": {}
}
\end{lstlisting}}
        & The model notices a parked car but decides to keep running toward the distant goal. It never attempts to board a vehicle and remains on foot until the countdown expires. Its plan omits the transport stage needed to finish in time. \\
        \addlinespace[4pt]
        \textbf{Calibration}
        & \caseimage{\includegraphics[width=\linewidth]{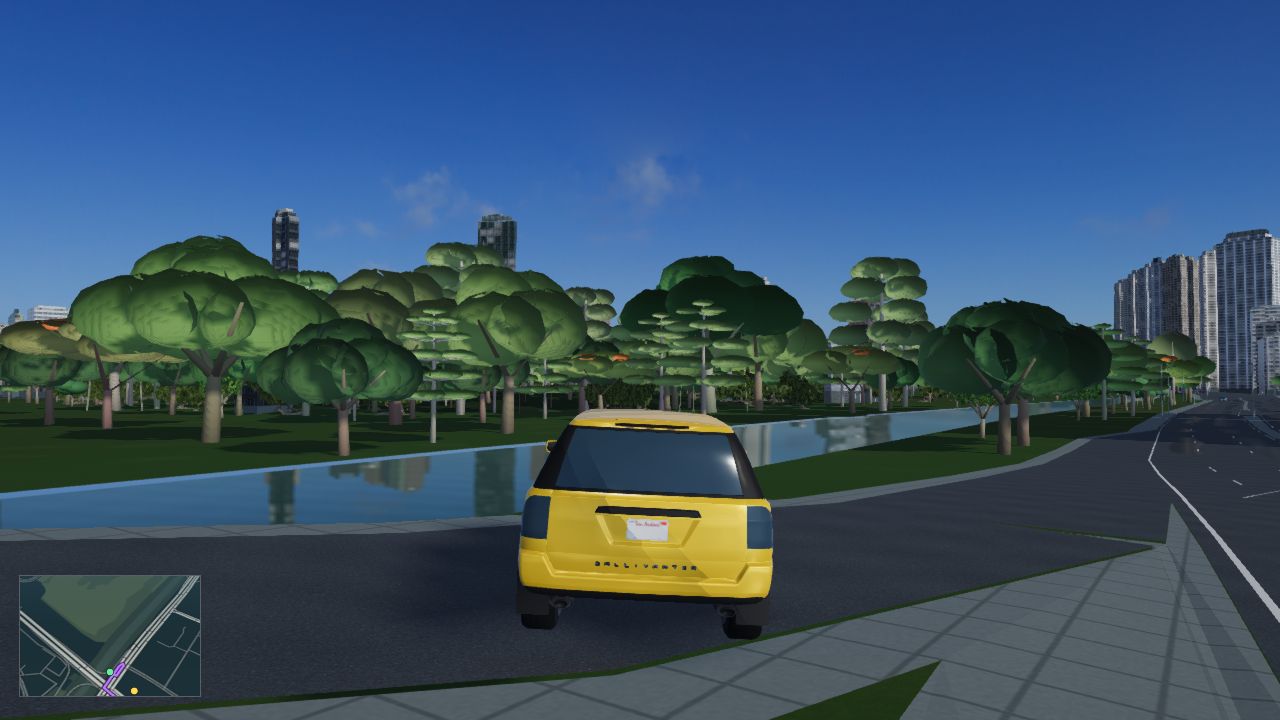}}{Kimi-K3}
        & \caseaction{
\begin{lstlisting}[language={},basicstyle=\ttfamily\fontsize{7}{8}\selectfont,
                columns=fullflexible,keepspaces=true,breaklines=true,breakatwhitespace=false,
                showstringspaces=false,upquote=true,frame=none,
                aboveskip=0pt,belowskip=0pt]
{
  "keys": [
    {"key": "w",
     "duration_seconds": 1},
    {"key": "d",
     "duration_seconds": 1}
  ],
  "mouse": {}
}
\end{lstlisting}}
        & While accelerating into the bend, the agent holds right steering for a full second. The prolonged input turns the vehicle too far, carrying it off the road onto the pavement and grass. The steering duration is too large for the required heading correction. \\
        \addlinespace[4pt]
        \textbf{Regulation}
        & \caseimage{\includegraphics[width=\linewidth]{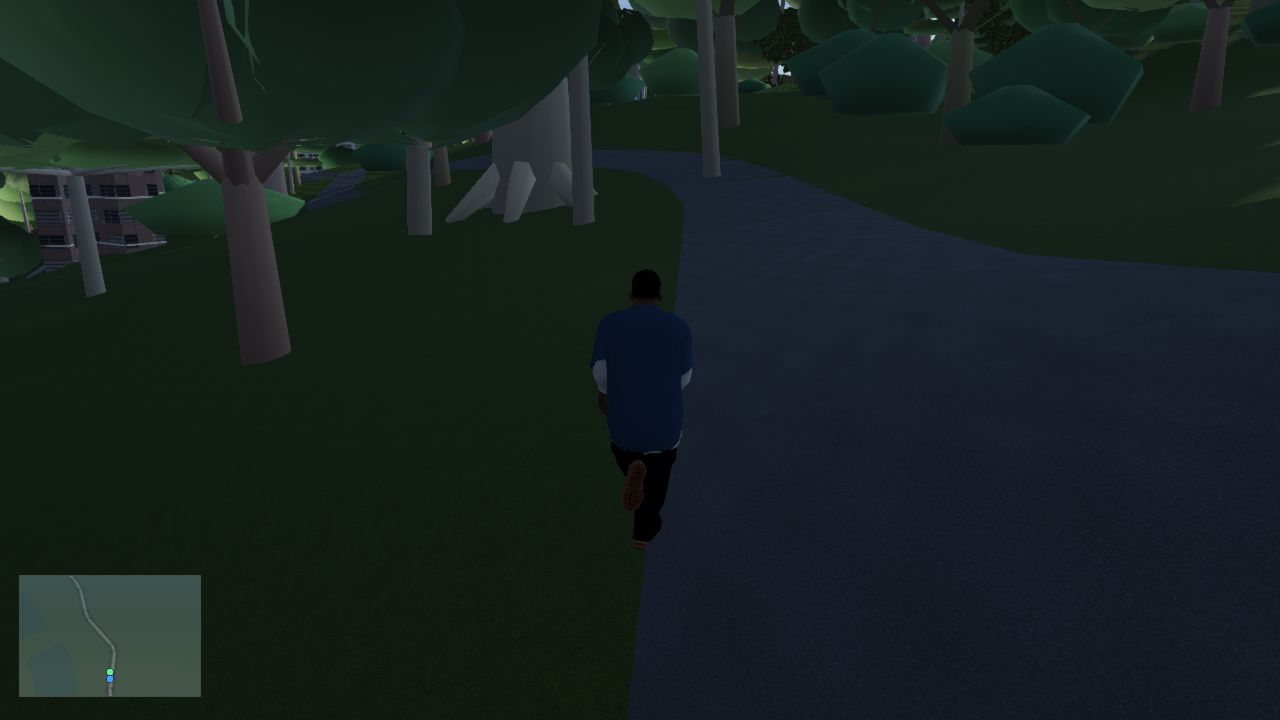}}{DeepSeek-V4.1-Flash}
        & \caseaction{
\begin{lstlisting}[language={},basicstyle=\ttfamily\fontsize{7}{8}\selectfont,
                columns=fullflexible,keepspaces=true,breaklines=true,breakatwhitespace=false,
                showstringspaces=false,upquote=true,frame=none,
                aboveskip=0pt,belowskip=0pt]
{
  "keys": [
    {"key": "w",
     "duration_seconds": 1},
    {"key": "Shift",
     "duration_seconds": 1}
  ],
  "mouse": {}
}
\end{lstlisting}}
        & The companion's route bends, but the agent keeps running straight without adjusting its heading. It moves off the road and repeats the input as the companion gap grows. The episode later fails when separation exceeds 20 meters. \\
        \bottomrule
    \end{tabular}
    \endgroup
    \caption{Representative error cases, one per reporting category. Screenshots precede the displayed actions. The Planning frame shows later evidence of a sustained missing-vehicle plan; the other frames use the annotated first-error observations.}
    \label{tab:first_error_cases}
\end{table*}

\FloatBarrier
\subsection{Recovery Criterion and Statistics}
\label{sec:appendix_error_recovery}
For a first error at decision $r$, recovery is assessed over the next ten decisions, $(r,r+10]$, truncated at the end of the recording.
Recovery requires either task completion or two consecutive later transitions demonstrating renewed task-relevant progress under the original constraints.

\paragraph{Outcomes and denominator.}
An error is \emph{recovered} if the criterion is met within the window, and \emph{not recovered} if the full window shows no recovery or an irreversible in-game failure occurs earlier.
If the recording or budget ends before recovery can be established, or the visual evidence remains inconclusive, the case is also counted as not recovered.
For model $m$ and capability $c$, the plotted rate is
\begin{equation}
    \rho_{m,c}=\frac{N^{\mathrm{rec}}_{m,c}}
    {N^{\mathrm{rec}}_{m,c}+N^{\mathrm{not\mbox{-}rec}}_{m,c}}.
\end{equation}
Here, $N^{\mathrm{rec}}_{m,c}+N^{\mathrm{not\mbox{-}rec}}_{m,c}=N^{\mathrm{err}}_{m,c}$ includes all categorized first-error cases for that model and capability, including those without conclusive recovery evidence.
Figure~\ref{fig:failure_recovery_heatmap} reports $100\rho_{m,c}$: recovery is conditional on an identified first error in that category, whereas first-error incidence uses all 150 tasks as its denominator.

\paragraph{Overall recovery.}
Among the 565 categorized cases, 179 recover and 386 are counted as not recovered.
The recovery rate is therefore 179/565 (31.7\%) over all categorized cases.
Using this same cohort, recovery is 65.7\% for Astra, 49.3\% for Sol, 22.2\% for DeepSeek, 24.0\% for GLM, 21.6\% for Kimi, and 25.6\% for Qwen.

Table~\ref{tab:first_error_recovery_rates} reports the numerical recovery rates for each model and capability, together with pooled rates.
\begin{table}[H]
    \centering
    \scriptsize
    \setlength{\tabcolsep}{2.5pt}
    \renewcommand{\arraystretch}{1.12}
    \begin{tabular}{@{}lrrrrr@{}}
        \toprule
        Model & Perception & Planning & Calibration & Regulation & Overall \\
        \midrule
        Astra & 20.0 & 100.0 & 69.2 & 67.6 & 65.7 \\
        Sol & 0.0 & 40.0 & 53.6 & 60.0 & 49.3 \\
        DeepSeek & 20.0 & 26.3 & 22.7 & 21.1 & 22.2 \\
        GLM & 29.2 & 21.4 & 25.9 & 19.4 & 24.0 \\
        Kimi & 8.0 & 33.3 & 20.7 & 25.6 & 21.6 \\
        Qwen-27B & 26.7 & 28.9 & 24.0 & 21.4 & 25.6 \\
        \midrule
        Pooled & 19.6 & 30.2 & 36.3 & 35.5 & 31.7 \\
        \bottomrule
    \end{tabular}
    \caption{Recovery rates (\%) within ten decisions, corresponding to Figure~\ref{fig:failure_recovery_heatmap}. All 565 categorized first errors are included; truncated and inconclusive cases count as not recovered. Overall and pooled rates use summed counts rather than averages of percentages.}
    \label{tab:first_error_recovery_rates}
\end{table}

\paragraph{Recovery examples.}
Table~\ref{tab:recovery_cases} illustrates the error state, corrective action, and restored progress for each capability.
% Generated by scripts/export_recovery_examples.py; sources in figures/recovery-cases/manifest.json.
\begin{table*}[p]
\centering
\footnotesize
\setlength{\tabcolsep}{3pt}
\begingroup
\newcommand{\recoverycell}[3]{\begin{minipage}[t]{\linewidth}\vspace{0pt}#1\par\vspace{2pt}\textbf{#2}\par\scriptsize #3\end{minipage}}
\begin{tabular}{@{}>{\raggedright\arraybackslash}p{0.12\textwidth}*{3}{>{\raggedright\arraybackslash}p{0.28\textwidth}}@{}}
\toprule
Capability / model & Error state $\longrightarrow$ & Corrective action $\longrightarrow$ & Restored progress \\
\midrule
\textbf{Perception}\par\scriptsize Qwen3.8-27B & \recoverycell{\includegraphics[width=\linewidth]{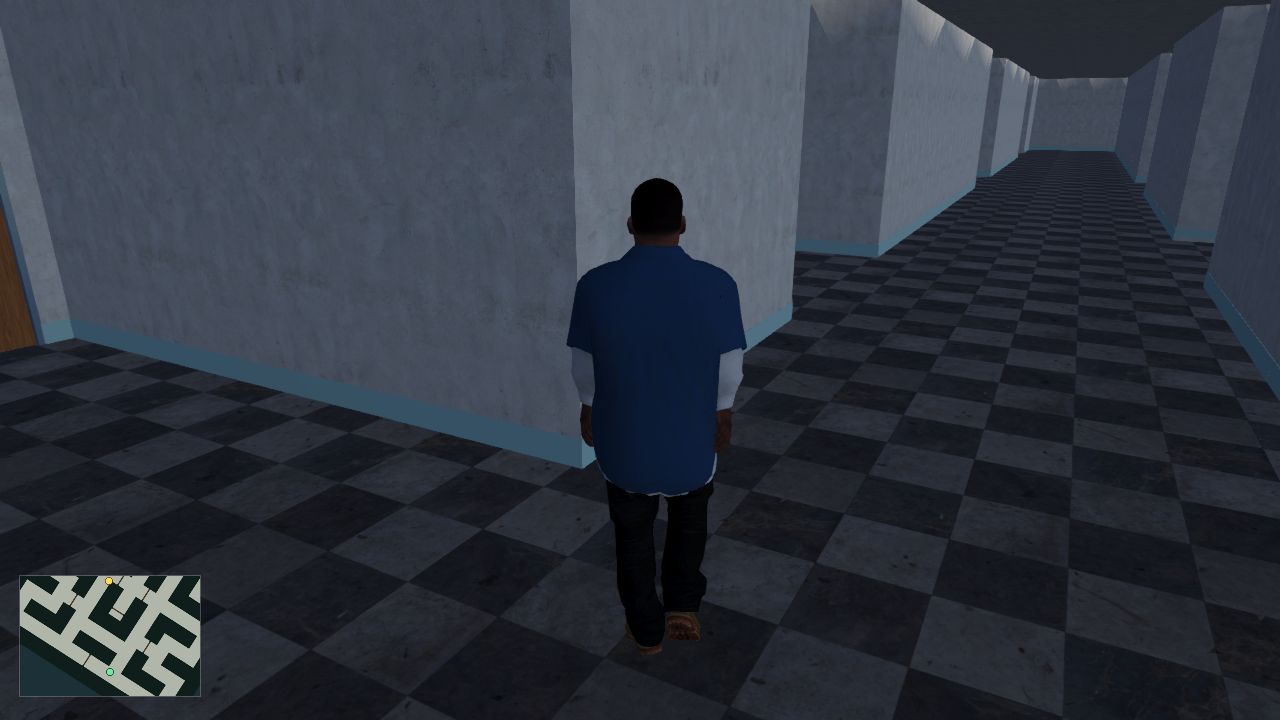}}{Confuse corridor direction}{After action: mouse (-250, 0)} & \recoverycell{\includegraphics[width=\linewidth]{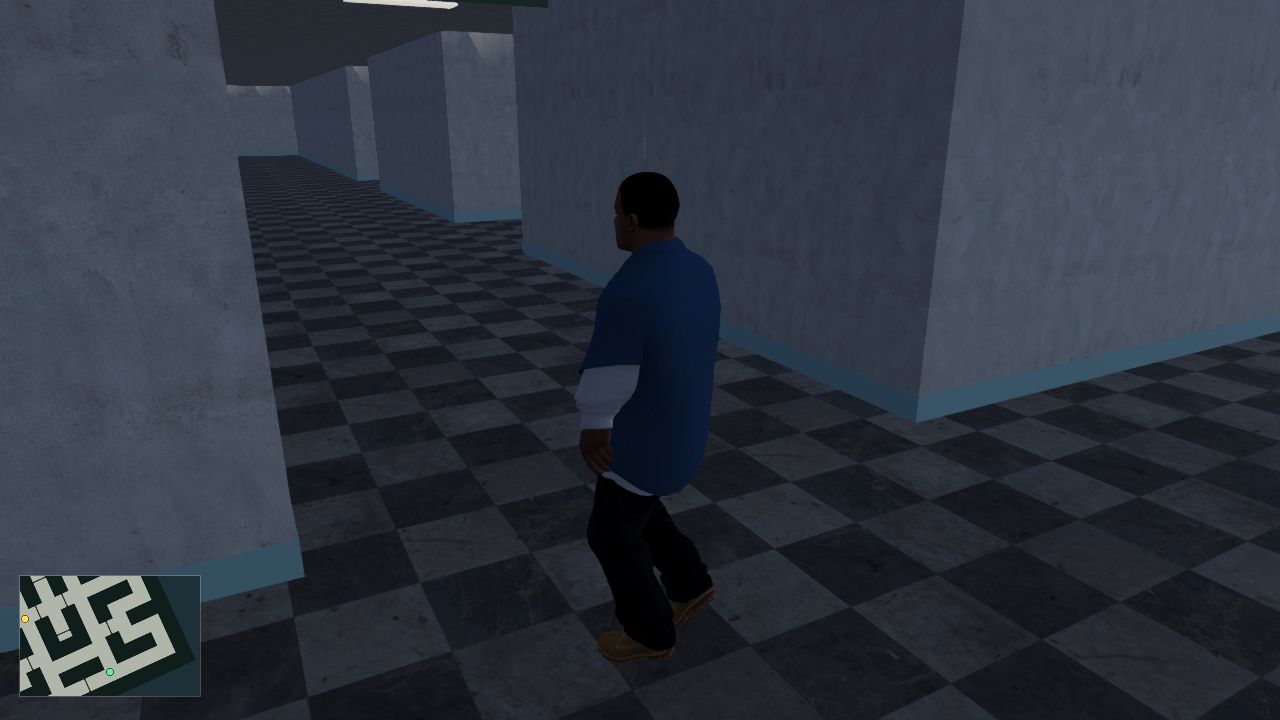}}{Recognize the alternating views}{Before action: W 1s} & \recoverycell{\includegraphics[width=\linewidth]{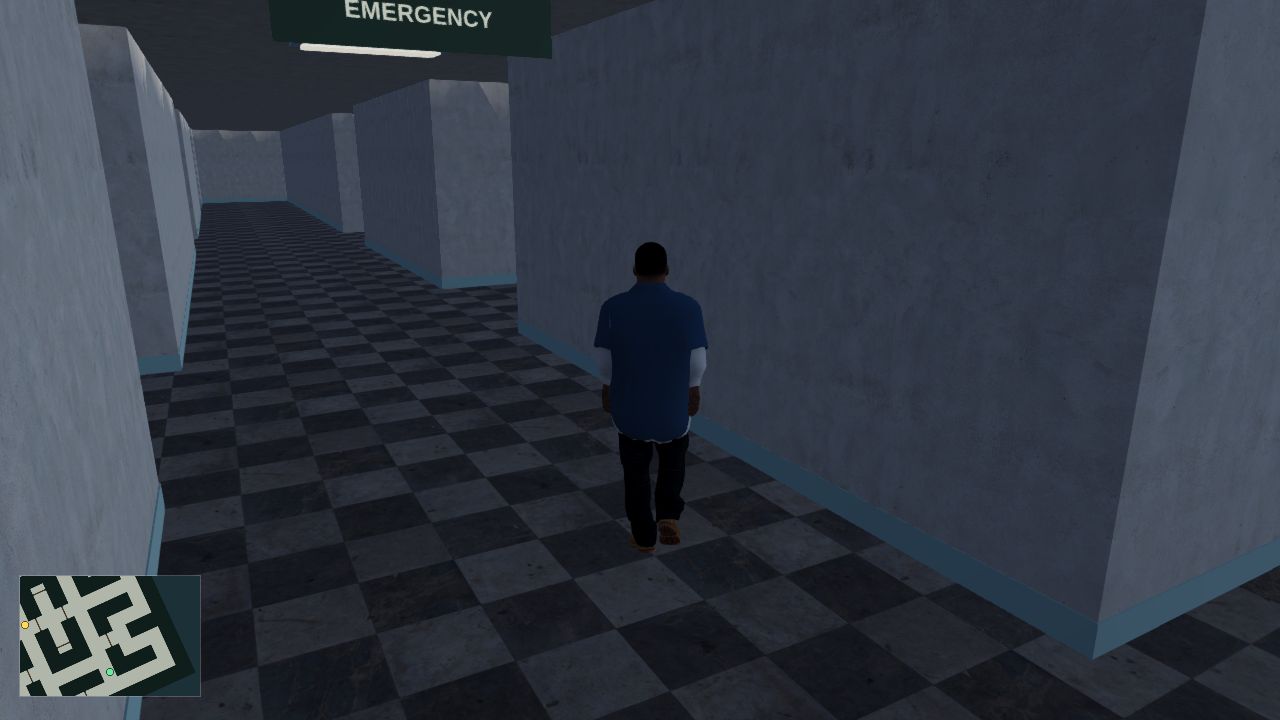}}{Advance along the corridor}{After action: W 1s} \\
 & \multicolumn{3}{p{0.85\textwidth}}{\footnotesize The agent alternates left/right views while inconsistently interpreting the corridor relative to the character. It then recognizes the oscillation, stops rotating the camera, and makes two consecutive forward moves along the corridor.} \\
\addlinespace[6pt]
\textbf{Planning}\par\scriptsize GPT-6-Astra & \recoverycell{\includegraphics[width=\linewidth]{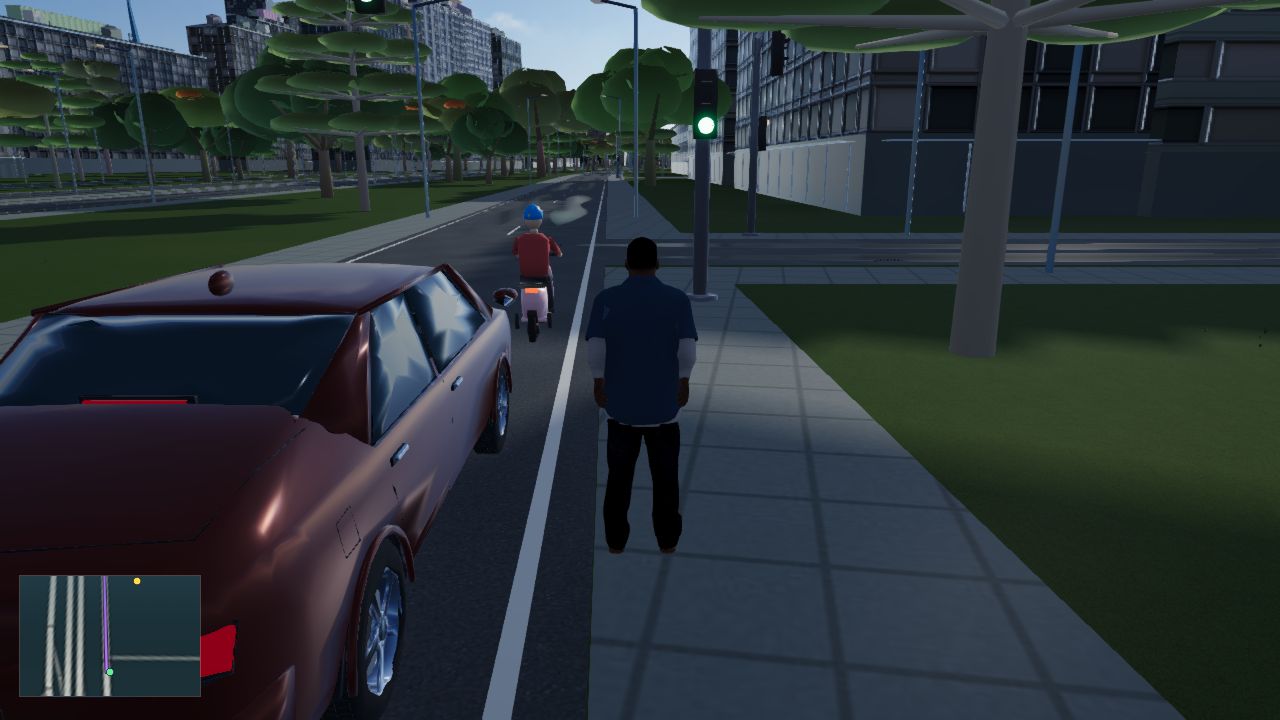}}{Leave the required vehicle}{After action: F 1s} & \recoverycell{\includegraphics[width=\linewidth]{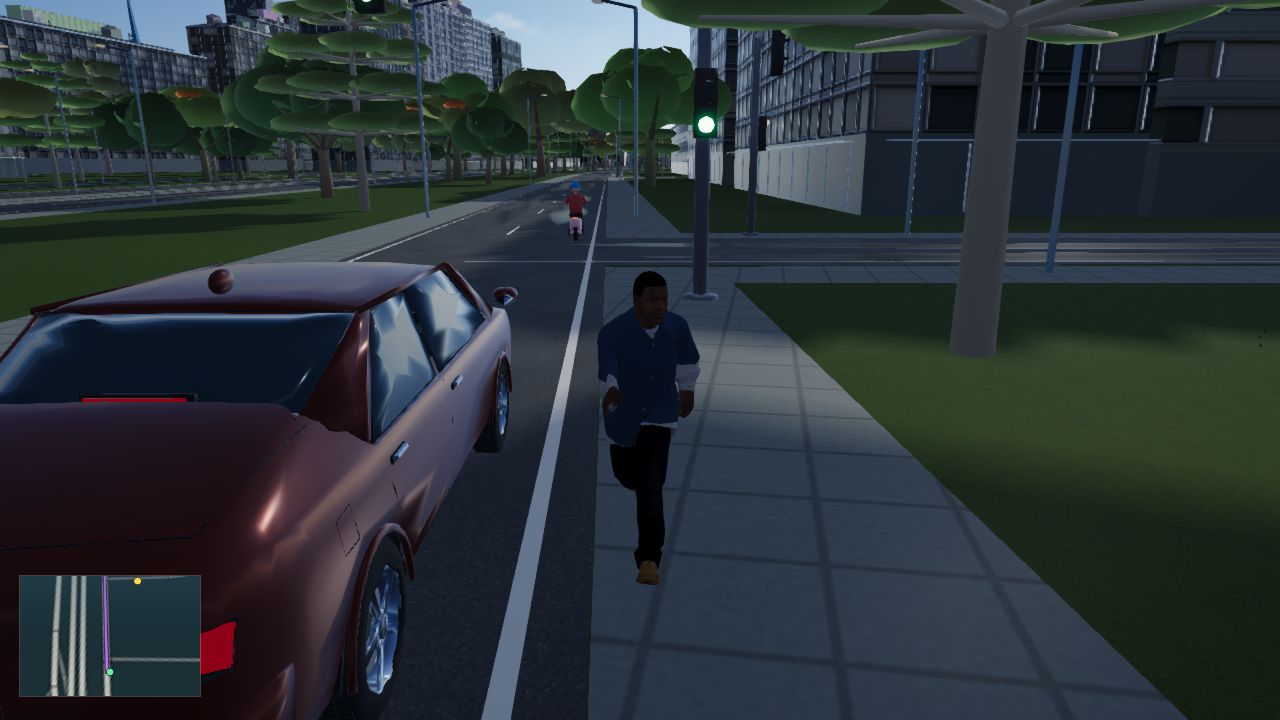}}{Return and request boarding}{Before action: F 1s} & \recoverycell{\includegraphics[width=\linewidth]{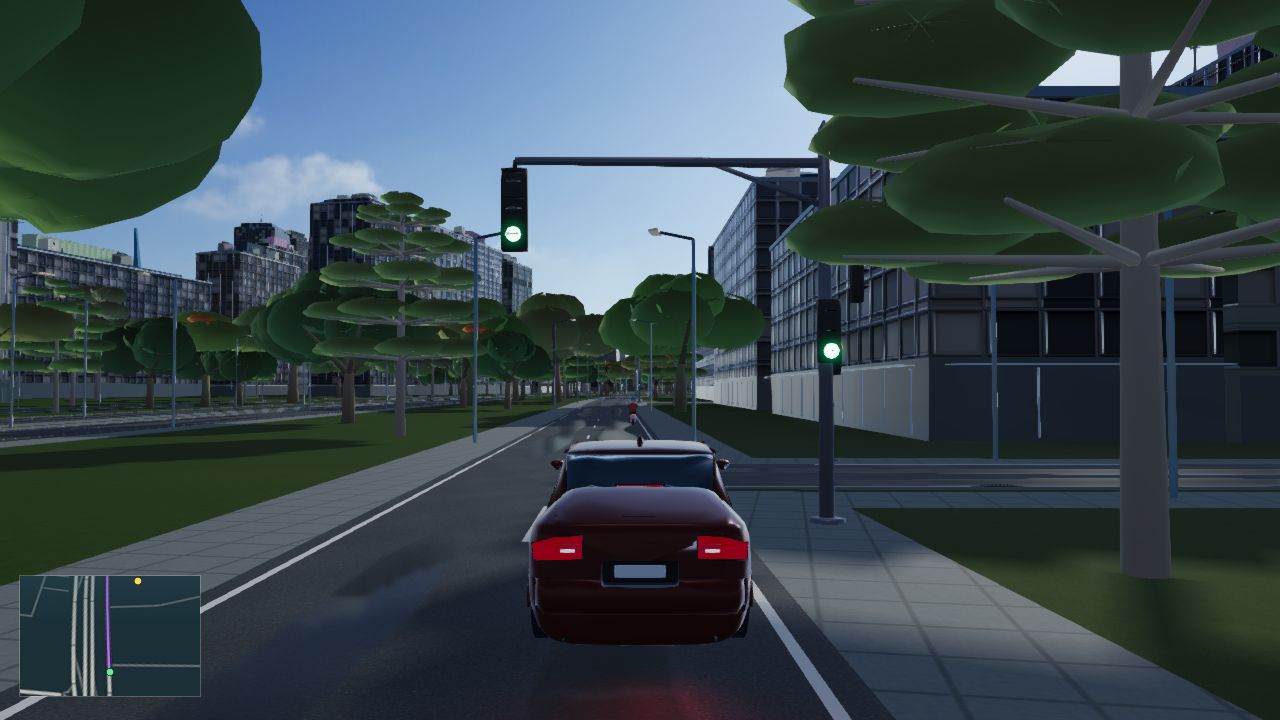}}{Resume driving toward the goal}{After action: W 1s} \\
 & \multicolumn{3}{p{0.85\textwidth}}{\footnotesize Exiting the car interrupts a task that requires driving. The agent returns to the vehicle, boards again, and resumes forward travel.} \\
\addlinespace[6pt]
\textbf{Calibration}\par\scriptsize GPT-6-Astra & \recoverycell{\includegraphics[width=\linewidth]{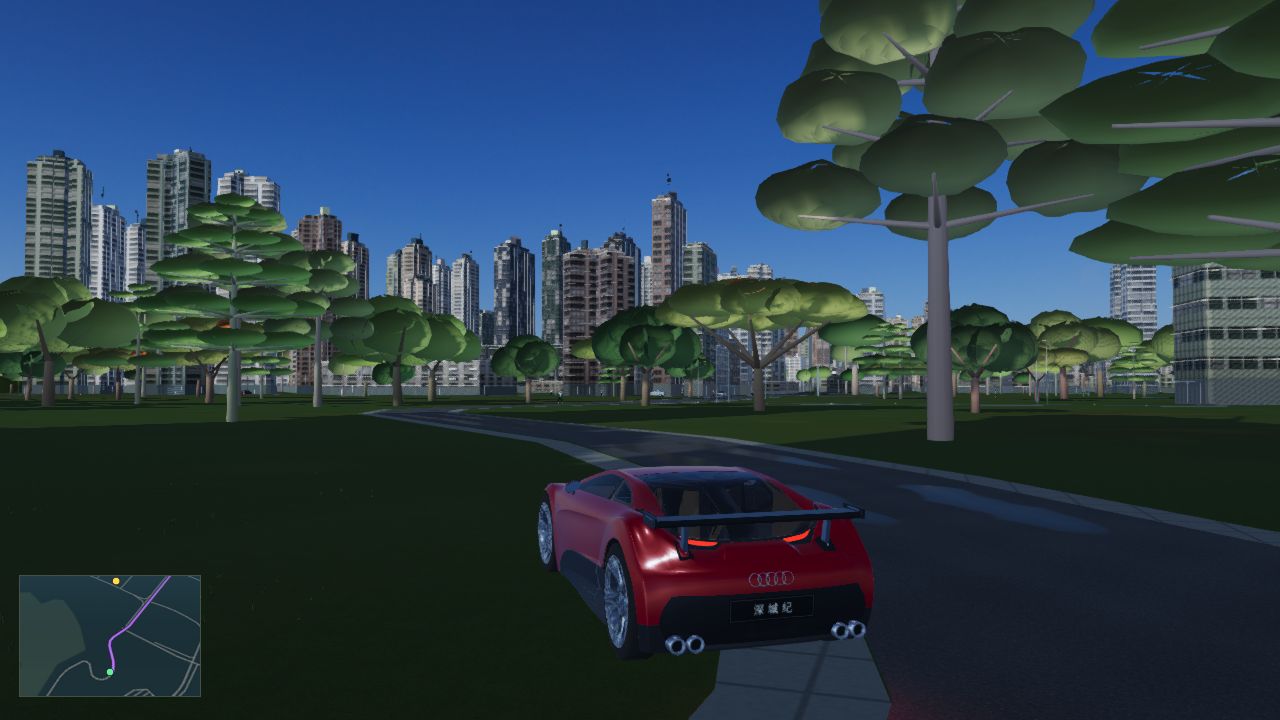}}{Oversteer onto the verge}{After action: W 1s; A 0.8s} & \recoverycell{\includegraphics[width=\linewidth]{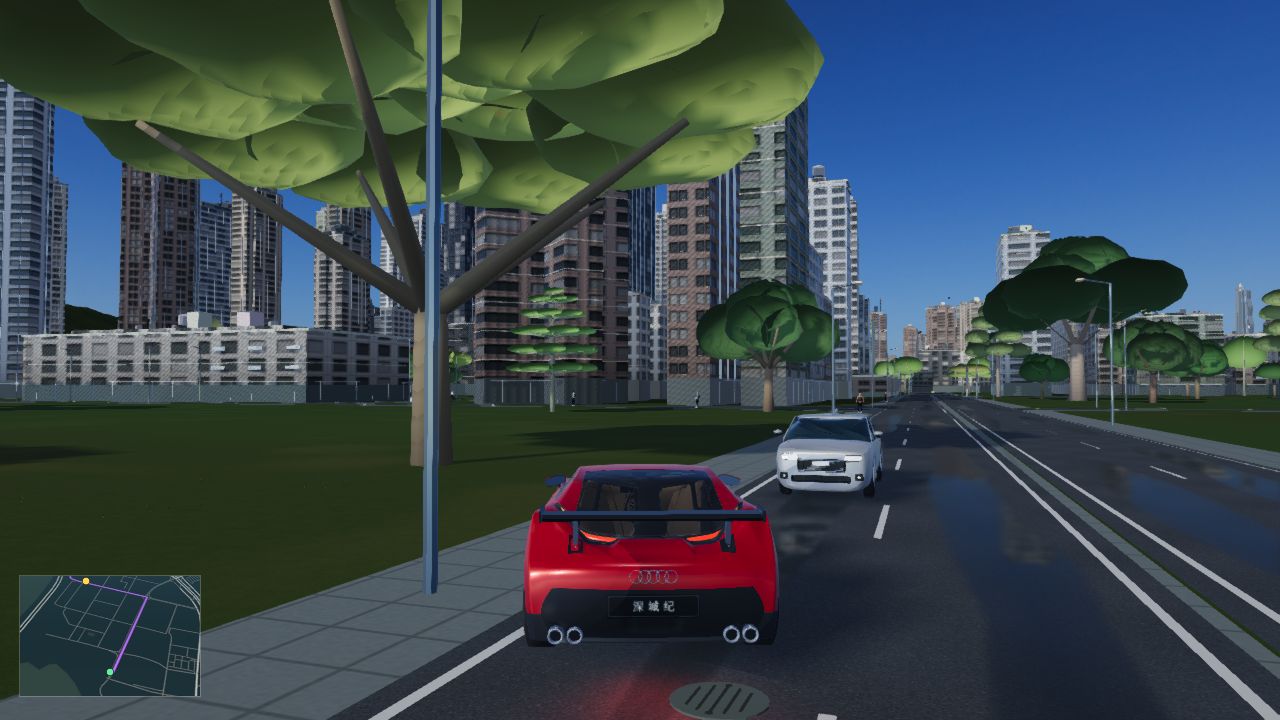}}{Apply a shorter steering correction}{Before action: W 1s; D 0.18s} & \recoverycell{\includegraphics[width=\linewidth]{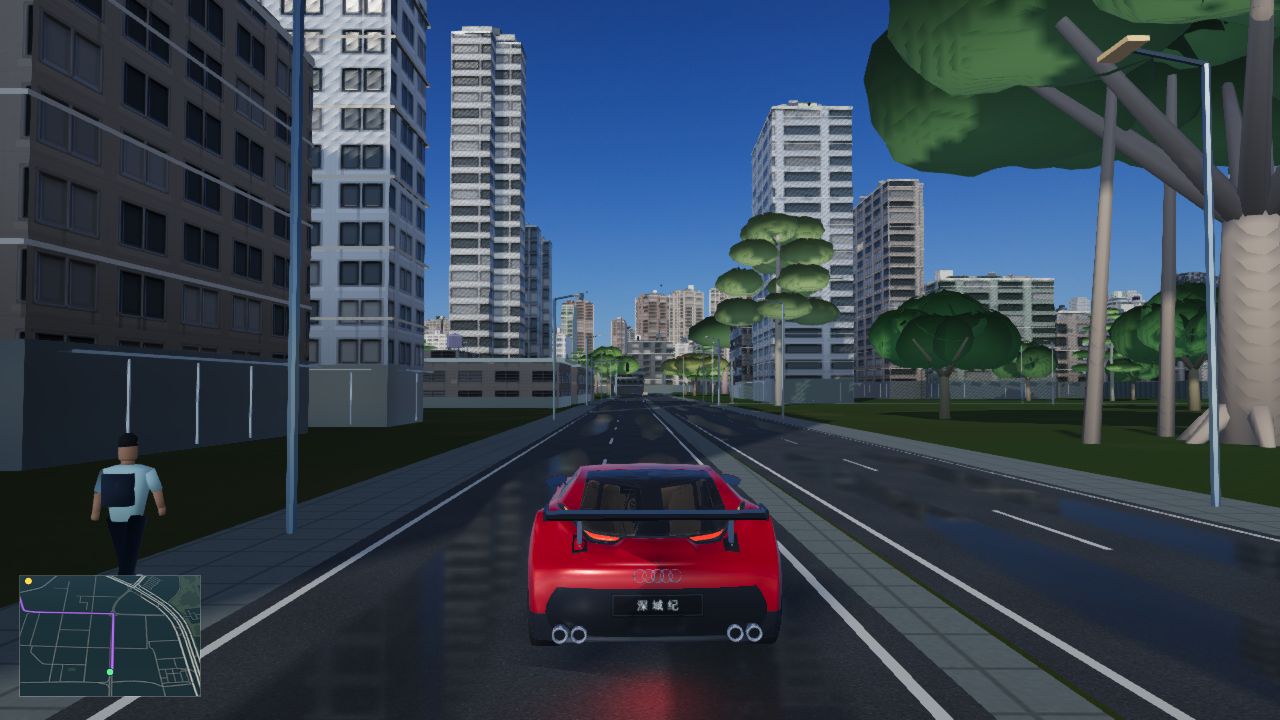}}{Continue within the road}{After action: W 1s; A 0.17s} \\
 & \multicolumn{3}{p{0.85\textwidth}}{\footnotesize A 0.8-second left input carries the car off the road. Subsequent shorter steering inputs bring it back into the lane and restore forward progress.} \\
\addlinespace[6pt]
\textbf{Regulation}\par\scriptsize GPT-6-Astra & \recoverycell{\includegraphics[width=\linewidth]{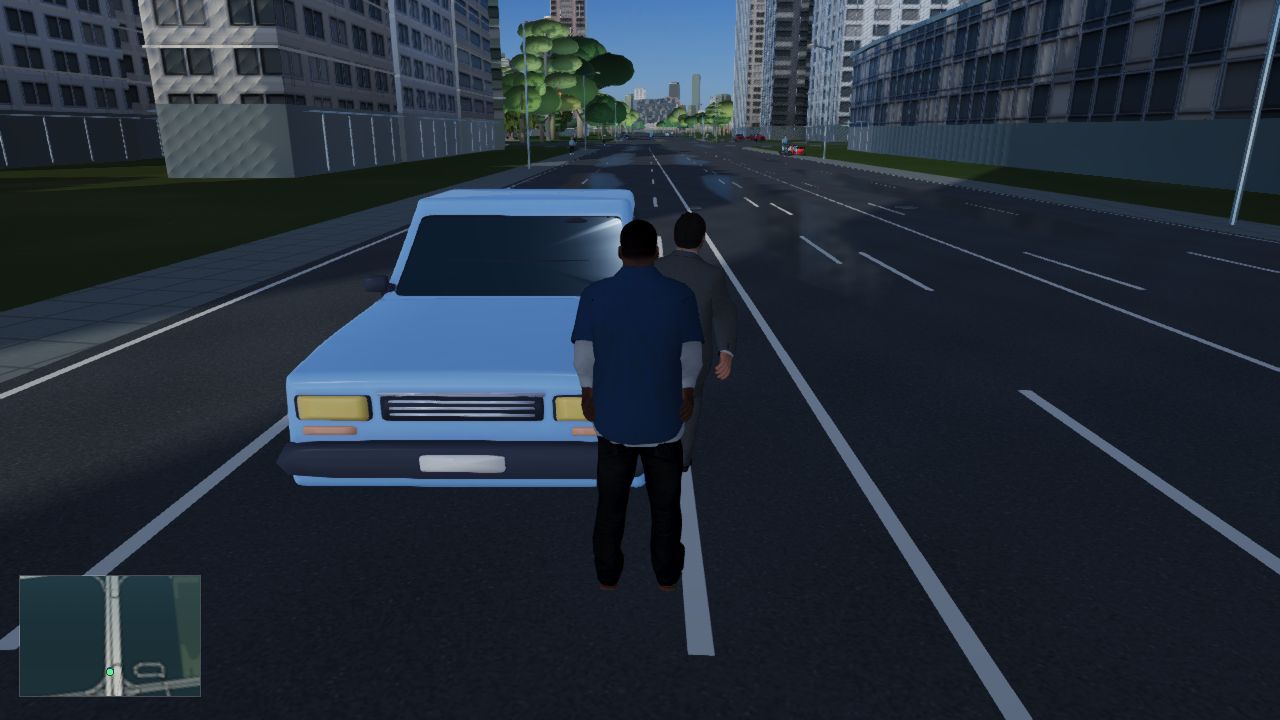}}{Blocked while following}{After action: W 1s} & \recoverycell{\includegraphics[width=\linewidth]{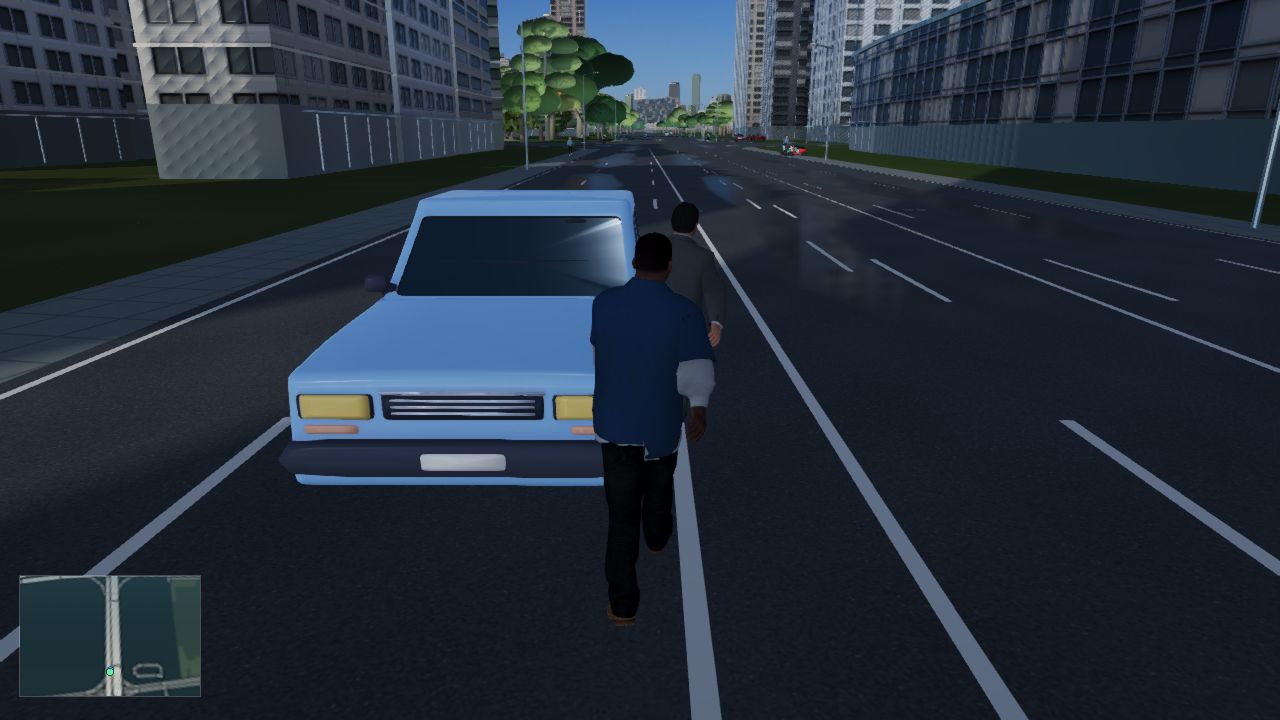}}{Sidestep around the vehicle}{Before action: D 0.8s; W 1s} & \recoverycell{\includegraphics[width=\linewidth]{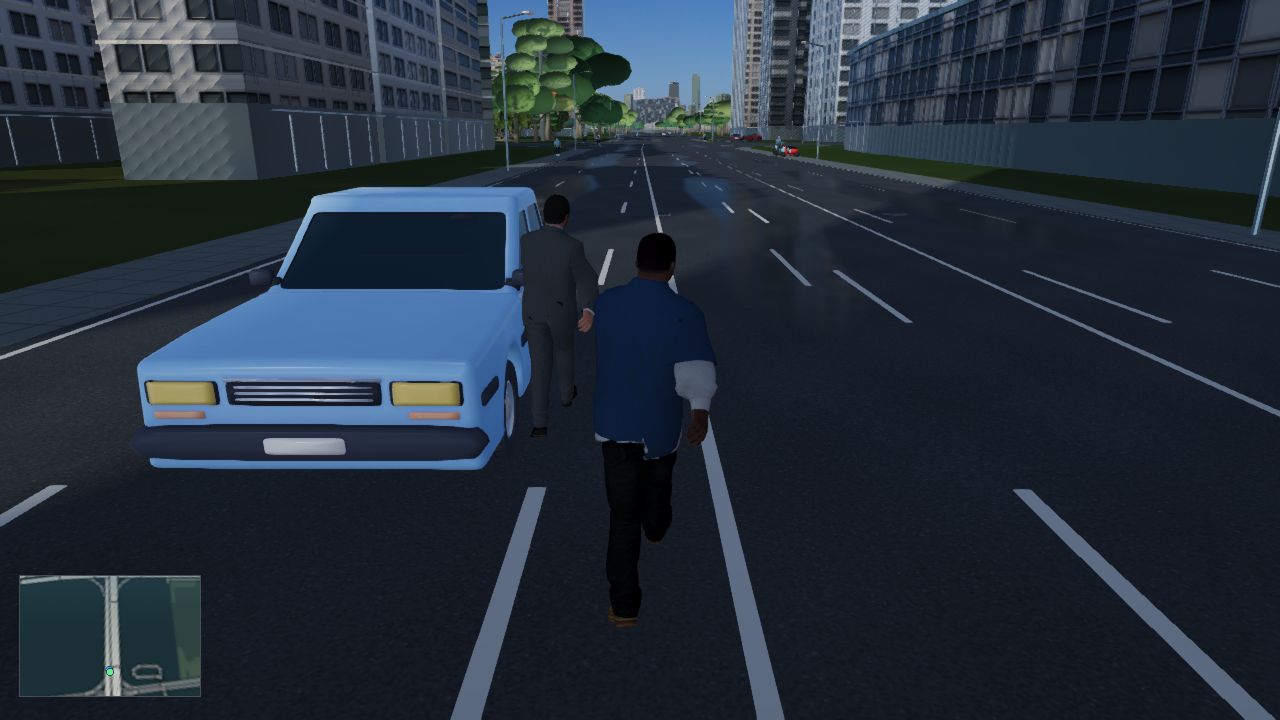}}{Resume following the companion}{After action: D 0.8s; W 1s} \\
 & \multicolumn{3}{p{0.85\textwidth}}{\footnotesize While following the companion, the agent continues forward into a blocking car and stops. It then adds rightward movement to clear the car, restoring consecutive forward progress with the companion still ahead.} \\
\addlinespace[6pt]
\bottomrule
\end{tabular}
\endgroup
\caption{Recovery sequences for the four capability categories. Annotations specify each observation's timing relative to the recorded action. Actions show key durations and mouse displacements; intermediate actions are omitted. The final observation follows the annotated recovery decision. Recovery satisfies the ten-decision criterion in Appendix~\ref{sec:appendix_error_recovery}: task completion or two consecutive transitions of renewed task progress.}
\label{tab:recovery_cases}
\end{table*}

% Keep prompt explanations with the preceding appendix text; full-width listings follow the example floats.
\FloatBarrier
\section{Evaluation Prompts}
\label{sec:appendix_prompts}
Listings~\ref{lst:gui_system_prompt} and~\ref{lst:codex_environment_prompt} reproduce our complete English prompt texts, including their Markdown headings and action examples.

\paragraph{\benchmark.}
Listing~\ref{lst:gui_system_prompt} is the entire shared system message.
The action adapter's schema description is not appended to it.
A task-specific instruction accompanies the first screenshot.
For online tasks, the host retains at most ten screenshot--reply pairs and drops the oldest five before the eleventh observation is submitted.
The instruction is attached to the earliest retained screenshot after truncation; the system message remains unchanged.
Image-based tasks retain their full interaction history.
Observation messages contain no additional round-number, elapsed-time, or cursor-coordinate text.

\paragraph{\benchmarkagi.}
Listing~\ref{lst:codex_environment_prompt} gives the environment instructions supplied through \texttt{AGENTS.md}, alongside Codex's native instructions.
It preserves the shared game controls while adapting action submission to \texttt{game\_step}.
The task objective appears only in the top-left corner of the initial screenshot; the initial user text is \texttt{play the game use game\_step}.
Subsequent tool results do not repeat the objective.
The harness retains and automatically compacts its native history, without the ten-image truncation rule used by \benchmark.

\paragraph{Game tool and prompt provenance.}
The model-facing tool description is reproduced in Listing~\ref{lst:game_step_description}.
Only \texttt{game\_step} is exposed for interacting with the environment, with a \texttt{request\_id} and an \texttt{action} object.
A successful action returns the next screenshot; rejected or stale calls return an error without an image.
The listed instructions and tool description come from an evaluation episode using Codex CLI 0.163.0.

\begingroup
% Review-mode line numbering interferes with breakable verbatim boxes.
\ifdefined\nolinenumbers\nolinenumbers\fi
\raggedbottom
\par\refstepcounter{promptlisting}\label{lst:gui_system_prompt}%
\begin{tcblisting}{    enhanced jigsaw,
    breakable,
    lines before break=4,
    listing only,
    listing options={style=evaluationprompt},
    colback=white,
    colbacktitle=black,
    colframe=gray!60,
    coltitle=white,
    fonttitle=\bfseries\small,
    title={Listing \thepromptlisting: \benchmark: complete system prompt},
    title after break={Listing \thepromptlisting: \benchmark: complete system prompt (continued)},
    arc=2mm,
    boxrule=0.4pt,
    titlerule=0pt,
    left=8pt,right=8pt,top=6pt,bottom=6pt,
    toptitle=3pt,bottomtitle=3pt,
    before skip=8pt,after skip=10pt}
You are a game-playing agent controlling GTA V with a keyboard and mouse. Complete the current task by controlling the character or vehicle using only screenshots from this game and the action history.

## Minimap, camera, and general controls

- On the bottom-left minimap, the player is a green dot. Static destinations are usually yellow dots, moving chase or follow targets are blue dots, and police or pursuers are red dots. Red sectors indicate capture zones. A dot on the minimap boundary indicates the direction of an off-map target, not its distance.
- The purple line shows the navigation route from your current position to the current target. In chase and follow tasks, it leads to the target's current position, not along the target's entire future route. Navigation may change with your position or the task stage. Indoor tasks have no purple route; you must find your own way.
- Mouse movement changes the view. On foot or while swimming, looking in a different direction also changes the direction of subsequent forward movement. While driving a car or boat, or flying, mouse movement does not steer the vehicle.
- V switches camera views. Ordinary cars cycle through near, medium, and far third-person views and the cockpit view; other modes use their supported views. Driving tasks start in third person. In driving modes that support it, hold C to look behind and release it to return; C does not toggle camera modes.
- While driving an ordinary car, the camera smoothly recenters when the vehicle is moving and there has been no mouse input for a while.
- F enters or exits a car, boat, or aircraft, subject to position, speed, and state constraints. Approach the vehicle before using it. F can also be used to take over vehicles.

## Controls

### On-foot controls

- W moves forward, S backward, and A/D left/right. Use the mouse to adjust your direction. Hold Shift while moving to run.
- Space triggers a jump on ordinary ground. Near and facing a configured railing, low wall, or ledge entry, the same key triggers a vault or climb.
- While climbing, W climbs up and S climbs down. Release the up/down input to stop. Shift accelerates climbing up or sliding down.

### Vehicle driving controls

- W applies forward throttle. S brakes while moving forward; keep holding it to reverse. A/D steer. Space applies the handbrake or brake. Motorcycles use the same controls.
- F enters or exits the vehicle when allowed and safe.

### Water controls

- While swimming, WASD moves horizontally and Shift increases speed. Hold Control to dive and Space to ascend. Without vertical input, maintain your current depth; pressing both ascent and descent cancels the vertical input.
- Underwater tasks require the correct horizontal position and depth.
- While driving a boat, W moves forward, S brakes/reverses, A/D steer, and Space brakes. The mouse is mainly for looking around.
- To board a boat, first reach the surface, approach the correct boarding side, and press F when the boat is nearly stationary. If the task requires driving a boat ashore, drive it onto the shore slope and onto land; swimming ashore does not count.

### Aviation controls

- Fixed-wing aircraft: W increases throttle; S reduces throttle/slows down; A/D control the rudder. Numpad5 pitches up, Numpad8 pitches down, and Numpad4/Numpad6 bank left/right. G toggles retractable landing gear.
- Helicopters: W increases lift/ascends; S reduces lift/descends; A/D control yaw. Numpad8 tilts forward/pitches down, Numpad5 tilts backward/pitches up, and Numpad4/Numpad6 bank left/right.
- Use only the numeric keypad controls for flight attitude: Numpad5, Numpad8, Numpad4, and Numpad6 in action JSON. The mouse looks around without changing flight attitude.

## Task descriptions

### Navigation tasks

- Reach the specified destination.

### Follow tasks

- Stay within the required distance of the target NPC throughout the task, until the NPC reaches its destination, while maintaining the required driving or on-foot state.

### Interception tasks

- In ordinary chases, intercept and stop the target NPC before it reaches its destination. If the target reaches its destination, you fail.
- Motorcycle chases require knocking the target down with three separate impacts. Sustained contact and impacts while it is down do not count again; let it recover and separate before hitting it again.

### Escape tasks

- Escape tasks allow free movement and have no fixed route or destination.
- The total time limit is 180 seconds. Succeed by staying outside every pursuer's forward red scanning sector for 60 consecutive seconds within that limit. Even brief exposure resets the 60-second escape timer.
- Remaining inside the pursuers' scanning sectors continuously for three seconds results in capture. Leaving all scanning sectors resets the continuous capture timer.
- Pursuers spawn dynamically from different directions. The red minimap sectors are the main risk indicator.

### Aviation tasks

- Aviation tasks may require ordered waypoints and specific altitudes. Apparent proximity on the two-dimensional minimap does not imply the correct altitude. Do not skip intermediate waypoints and fly straight to the final landing site.
- Hover tasks generally require staying within about 12 meters of the target circle at no more than 3 m/s for three consecutive seconds. Leaving these conditions resets the timer.
- After completing the aerial objectives, follow the current approach and landing stage. For fixed-wing aircraft, align with the assigned runway, lower the landing gear, touch down, and use S to reduce throttle and stop. For helicopters, slow over the designated H helipad and descend gently with S. Reduce throttle and remain stationary for two seconds inside the valid landing area.

### Shooting and sniping tasks

These are first-person tasks. The input image is 1280x720, and the aiming center is fixed at (640,360). Move the mouse to change the view, align the instructed target with the center, and press the left mouse button to fire. Only sniping requires using the scope and adjusting magnification; ordinary shooting does not. For ordinary shooting, move to aim and then fire. In sniping, there is no center circle while unscoped, and mouse movement has no effect and does not accumulate. Press the right mouse button to open the scope, adjust your aim or scroll to change magnification in subsequent turns, and finally press the left mouse button to fire. Closing the scope disables movement again. No numeric magnification indicator is displayed. Each turn allows only one operation: aim movement, opening the scope, zooming, or firing.

## Action interface and output format

All task screenshots are **1280x720 pixels**.

All tasks use pixel-based relative mouse displacement `mouse.move:{dx,dy}`: **positive dx moves right, negative dx left; positive dy moves up, negative dy down**. Each turn adjusts the current view; repeating a nonzero delta adjusts it again. `{dx:0,dy:0}` does not move. The aiming center in shooting and sniping remains **(640,360)**. Native 3D rotates using camera sensitivity and does not guarantee exact centering of a selected object.

Control the game through the unified action interface. Each reply must be one complete JSON object with only the top-level fields `keys`, `mouse`, and `wait_seconds`. Output nothing other than JSON. `keys` is required; use `[]` when there is no keyboard input. `mouse` is optional; `{}` means no mouse operation.

`keys` specifies keyboard input. Within `mouse`, `move`, `scroll`, and `buttons` specify movement, scrolling, and button presses/releases that persist across turns. Only keyboard input has `duration_seconds`; mouse movement, scrolling, and mouse buttons have no duration field.

Interface rules:

- `keys` is a required array. Each item contains only `key` and `duration_seconds`. Standard key names are `w`, `a`, `s`, `d`, `Shift`, `Control`, `Space`, `e`, `f`, `q`, `c`, `v`, `g`, `Numpad4`, `Numpad5`, `Numpad6`, and `Numpad8`. Each key has an independent hold duration. Do not repeat a key. `duration_seconds` must be a finite number greater than zero and at most one, measured in game seconds, not a string. Durations above one second are rejected.
- With keyboard input, a turn's duration is automatically set to the maximum of its keyboard hold durations.
- Keys held through the end of a turn retain their state. Including the same key in the next turn continues the hold.
- Mouse movement is instantaneous at the beginning of a turn and consumes no game time. Keyboard holds then start together or continue from the previous turn. Keys whose durations end before the turn ends are released at their own deadlines.
- `mouse` may contain `buttons` and at most one of `move:{dx,dy}` or `scroll:{direction:"up" or "down",steps:1}`, never both. `scroll.steps` must be an integer from 1 to 6.
- `move` contains only numeric `dx` and `dy`, specifying relative pixel displacement, not an absolute position or normalized coordinates. **-640 <= dx <= 640 and -360 <= dy <= 360**; finite fractional values are allowed. Movement occurs once instantly, without continued turning during keyboard holds.
- `mouse.buttons` is an optional array. Each item contains only `button` (`left` or `right`) and `event` (`down` or `up`), with at most one event per button per turn. `down` presses and holds; `up` releases. Held mouse buttons persist across turns without repeated presses.

- Shooting and sniping tasks require no keyboard input: use `keys:[]`.
- Ordinary shooting ignores well-formed right-button and scroll operations, without opening a scope or zooming. If there are no other effective mouse operations, the reply is a no-op that still consumes one turn. Right-button and scroll controls remain functional in sniping tasks.
- `wait_seconds` is an optional finite number greater than zero and at most one, allowed only with `keys:[]`. In native 3D tasks, optional instantaneous mouse operations happen first, then keyboard keys are released and the game advances for the specified time. Mouse buttons retain their cross-turn hold rules; explicitly send `up` to release them. Shooting and sniping are discrete tasks: waiting advances no scene time but still consumes one turn.

## Operation examples

Advance the native game for 0.5 seconds without keyboard input or mouse movement (already-held mouse buttons remain held):

```json
{"keys": [], "mouse": {}, "wait_seconds": 0.5}
```

While on foot, hold W and Shift together to run for one second, first moving the view 120 pixels right and 30 pixels up at the start of the turn:

```json
{"keys": [{"key": "w", "duration_seconds": 1}, {"key": "Shift", "duration_seconds": 1}], "mouse": {"move": {"dx": 120, "dy": 30}}}
```

Press and hold the right mouse button to open the sniper scope:

```json
{"keys": [], "mouse": {"buttons": [{"button": "right", "event": "down"}]}}
```

With the scope already open, scroll up two notches in the next turn to increase magnification:

```json
{"keys": [], "mouse": {"scroll": {"direction": "up", "steps": 2}}}
```

In shooting or sniping, bring (700,340) in the current screenshot to the center crosshair (60 pixels right and 20 up; sniping requires an open scope, and scene-boundary limits apply):

```json
{"keys": [], "mouse": {"move": {"dx": 60, "dy": 20}}}
```
\end{tcblisting}

\par\refstepcounter{promptlisting}\label{lst:codex_environment_prompt}%
\begin{tcblisting}{    enhanced jigsaw,
    breakable,
    lines before break=4,
    listing only,
    listing options={style=evaluationprompt},
    colback=white,
    colbacktitle=black,
    colframe=gray!60,
    coltitle=white,
    fonttitle=\bfseries\small,
    title={Listing \thepromptlisting: \benchmarkagi: complete AGENTS.md environment instructions},
    title after break={Listing \thepromptlisting: \benchmarkagi: complete AGENTS.md environment instructions (continued)},
    arc=2mm,
    boxrule=0.4pt,
    titlerule=0pt,
    left=8pt,right=8pt,top=6pt,bottom=6pt,
    toptitle=3pt,bottomtitle=3pt,
    before skip=8pt,after skip=10pt}
You are a game-playing agent controlling GTA V with a keyboard and mouse. Complete the current task by controlling the character or vehicle using only screenshots from this game and the action history. The task instruction appears only in the top-left of the initial screenshot supplied by the evaluator in the initial user message. Subsequent action results do not repeat it.

## Minimap, camera, and general controls

- On the bottom-left minimap, the player is a green dot. Static destinations are usually yellow dots, moving chase or follow targets are blue dots, and police or pursuers are red dots. Red sectors indicate capture zones. A dot on the minimap boundary indicates the direction of an off-map target, not its distance.
- The purple line shows the navigation route from your current position to the current target. In chase and follow tasks, it leads to the target's current position, not along the target's entire future route. Navigation may change with your position or the task stage. Indoor tasks have no purple route; you must find your own way.
- Mouse movement changes the view. On foot or while swimming, looking in a different direction also changes the direction of subsequent forward movement. While driving a car or boat, or flying, mouse movement does not steer the vehicle.
- V switches camera views. Ordinary cars cycle through near, medium, and far third-person views and the cockpit view; other modes use their supported views. Driving tasks start in third person. In driving modes that support it, hold C to look behind and release it to return; C does not toggle camera modes.
- While driving an ordinary car, the camera smoothly recenters when the vehicle is moving and there has been no mouse input for a while.
- F enters or exits a car, boat, or aircraft, subject to position, speed, and state constraints. Approach the vehicle before using it. F can also be used to take over vehicles.

## Controls

### On-foot controls

- W moves forward, S backward, and A/D left/right. Use the mouse to adjust your direction. Hold Shift while moving to run.
- Space triggers a jump on ordinary ground. Near and facing a configured railing, low wall, or ledge entry, the same key triggers a vault or climb.
- While climbing, W climbs up and S climbs down. Release the up/down input to stop. Shift accelerates climbing up or sliding down.

### Vehicle driving controls

- W applies forward throttle. S brakes while moving forward; keep holding it to reverse. A/D steer. Space applies the handbrake or brake. Motorcycles use the same controls.
- F enters or exits the vehicle when allowed and safe.

### Water controls

- While swimming, WASD moves horizontally and Shift increases speed. Hold Control to dive and Space to ascend. Without vertical input, maintain your current depth; pressing both ascent and descent cancels the vertical input.
- Underwater tasks require the correct horizontal position and depth.
- While driving a boat, W moves forward, S brakes/reverses, A/D steer, and Space brakes. The mouse is mainly for looking around.
- To board a boat, first reach the surface, approach the correct boarding side, and press F when the boat is nearly stationary. If the task requires driving a boat ashore, drive it onto the shore slope and onto land; swimming ashore does not count.

### Aviation controls

- Fixed-wing aircraft: W increases throttle; S reduces throttle/slows down; A/D control the rudder. Numpad5 pitches up, Numpad8 pitches down, and Numpad4/Numpad6 bank left/right. G toggles retractable landing gear.
- Helicopters: W increases lift/ascends; S reduces lift/descends; A/D control yaw. Numpad8 tilts forward/pitches down, Numpad5 tilts backward/pitches up, and Numpad4/Numpad6 bank left/right.
- Use only the numeric keypad controls for flight attitude: Numpad5, Numpad8, Numpad4, and Numpad6 in action JSON. The mouse looks around without changing flight attitude.

## Task descriptions

### Navigation tasks

- Reach the specified destination.

### Follow tasks

- Stay within the required distance of the target NPC throughout the task, until the NPC reaches its destination, while maintaining the required driving or on-foot state.

### Interception tasks

- In ordinary chases, intercept and stop the target NPC before it reaches its destination. If the target reaches its destination, you fail.
- Motorcycle chases require knocking the target down with three separate impacts. Sustained contact and impacts while it is down do not count again; let it recover and separate before hitting it again.

### Escape tasks

- Escape tasks allow free movement and have no fixed route or destination.
- The total time limit is 180 seconds. Succeed by staying outside every pursuer's forward red scanning sector for 60 consecutive seconds within that limit. Even brief exposure resets the 60-second escape timer.
- Remaining inside the pursuers' scanning sectors continuously for three seconds results in capture. Leaving all scanning sectors resets the continuous capture timer.
- Pursuers spawn dynamically from different directions. The red minimap sectors are the main risk indicator.

### Aviation tasks

- Aviation tasks may require ordered waypoints and specific altitudes. Apparent proximity on the two-dimensional minimap does not imply the correct altitude. Do not skip intermediate waypoints and fly straight to the final landing site.
- Hover tasks generally require staying within about 12 meters of the target circle at no more than 3 m/s for three consecutive seconds. Leaving these conditions resets the timer.
- After completing the aerial objectives, follow the current approach and landing stage. For fixed-wing aircraft, align with the assigned runway, lower the landing gear, touch down, and use S to reduce throttle and stop. For helicopters, slow over the designated H helipad and descend gently with S. Reduce throttle and remain stationary for two seconds inside the valid landing area.

### Shooting and sniping tasks

These are first-person tasks. The input image is 1280x720, and the aiming center is fixed at (640,360). Move the mouse to change the view, align the instructed target with the center, and press the left mouse button to fire. Only sniping requires using the scope and adjusting magnification; ordinary shooting does not. For ordinary shooting, move to aim and then fire. In sniping, there is no center circle while unscoped, and mouse movement has no effect and does not accumulate. Press the right mouse button to open the scope, adjust your aim or scroll to change magnification in subsequent turns, and finally press the left mouse button to fire. Closing the scope disables movement again. No numeric magnification indicator is displayed. Each turn allows only one operation: aim movement, opening the scope, zooming, or firing.

## Action interface and output format

All task screenshots are **1280x720 pixels**.

All tasks use pixel-based relative mouse displacement `mouse.move:{dx,dy}`: **positive dx moves right, negative dx left; positive dy moves up, negative dy down**. Each turn adjusts the current view; repeating a nonzero delta adjusts it again. `{dx:0,dy:0}` does not move. The aiming center in shooting and sniping remains **(640,360)**. Native 3D rotates using camera sensitivity and does not guarantee exact centering of a selected object.

Control the game through the unified action interface. Submit the complete JSON object as `game_step.action`, together with a unique `request_id`; reuse the same `request_id` and identical action when retrying an uncertain call. The action object may contain only the top-level fields `keys`, `mouse`, and `wait_seconds`. `keys` is required; use `[]` when there is no keyboard input. `mouse` is optional; `{}` means no mouse operation.

`keys` specifies keyboard input. Within `mouse`, `move`, `scroll`, and `buttons` specify movement, scrolling, and button presses/releases that persist across turns. Only keyboard input has `duration_seconds`; mouse movement, scrolling, and mouse buttons have no duration field.

Interface rules:

- `keys` is a required array. Each item contains only `key` and `duration_seconds`. Standard key names are `w`, `a`, `s`, `d`, `Shift`, `Control`, `Space`, `e`, `f`, `q`, `c`, `v`, `g`, `Numpad4`, `Numpad5`, `Numpad6`, and `Numpad8`. Each key has an independent hold duration. Do not repeat a key. `duration_seconds` must be a finite number greater than zero and at most one, measured in game seconds, not a string. Durations above one second are rejected.
- With keyboard input, a turn's duration is automatically set to the maximum of its keyboard hold durations.
- Keys held through the end of a turn retain their state. Including the same key in the next turn continues the hold.
- Mouse movement is instantaneous at the beginning of a turn and consumes no game time. Keyboard holds then start together or continue from the previous turn. Keys whose durations end before the turn ends are released at their own deadlines.
- `mouse` may contain `buttons` and at most one of `move:{dx,dy}` or `scroll:{direction:"up" or "down",steps:1}`, never both. `scroll.steps` must be an integer from 1 to 6.
- `move` contains only numeric `dx` and `dy`, specifying relative pixel displacement, not an absolute position or normalized coordinates. **-640 <= dx <= 640 and -360 <= dy <= 360**; finite fractional values are allowed. Movement occurs once instantly, without continued turning during keyboard holds.
- `mouse.buttons` is an optional array. Each item contains only `button` (`left` or `right`) and `event` (`down` or `up`), with at most one event per button per turn. `down` presses and holds; `up` releases. Held mouse buttons persist across turns without repeated presses.

- Shooting and sniping tasks require no keyboard input: use `keys:[]`.
- Ordinary shooting ignores well-formed right-button and scroll operations, without opening a scope or zooming. If there are no other effective mouse operations, the reply is a no-op that still consumes one turn. Right-button and scroll controls remain functional in sniping tasks.
- `wait_seconds` is an optional finite number greater than zero and at most one, allowed only with `keys:[]`. In native 3D tasks, optional instantaneous mouse operations happen first, then keyboard keys are released and the game advances for the specified time. Mouse buttons retain their cross-turn hold rules; explicitly send `up` to release them. Shooting and sniping are discrete tasks: waiting advances no scene time but still consumes one turn.

## Operation examples

Advance the native game for 0.5 seconds without keyboard input or mouse movement (already-held mouse buttons remain held):

```json
{"keys": [], "mouse": {}, "wait_seconds": 0.5}
```

While on foot, hold W and Shift together to run for one second, first moving the view 120 pixels right and 30 pixels up at the start of the turn:

```json
{"keys": [{"key": "w", "duration_seconds": 1}, {"key": "Shift", "duration_seconds": 1}], "mouse": {"move": {"dx": 120, "dy": 30}}}
```

Press and hold the right mouse button to open the sniper scope:

```json
{"keys": [], "mouse": {"buttons": [{"button": "right", "event": "down"}]}}
```

With the scope already open, scroll up two notches in the next turn to increase magnification:

```json
{"keys": [], "mouse": {"scroll": {"direction": "up", "steps": 2}}}
```

In shooting or sniping, bring (700,340) in the current screenshot to the center crosshair (60 pixels right and 20 up; sniping requires an open scope, and scene-boundary limits apply):

```json
{"keys": [], "mouse": {"move": {"dx": 60, "dy": 20}}}
```
\end{tcblisting}

\par\refstepcounter{promptlisting}\label{lst:game_step_description}%
\begin{tcblisting}{    enhanced jigsaw,
    breakable,
    lines before break=4,
    listing only,
    listing options={style=evaluationprompt},
    colback=white,
    colbacktitle=black,
    colframe=gray!60,
    coltitle=white,
    fonttitle=\bfseries\small,
    title={Listing \thepromptlisting: \benchmarkagi: model-facing game\_step tool description},
    title after break={Listing \thepromptlisting: \benchmarkagi: model-facing game\_step tool description (continued)},
    arc=2mm,
    boxrule=0.4pt,
    titlerule=0pt,
    left=8pt,right=8pt,top=6pt,bottom=6pt,
    toptitle=3pt,bottomtitle=3pt,
    before skip=8pt,after skip=10pt}
Execute exactly one game action and return the next screenshot.
Key holds start together, last independently for at most 1 game second, and may continue across calls.
Mouse move is instantaneous relative pixels (dx positive right, dy positive up); mouse button holds persist until up.
Move and scroll cannot coexist.
wait_seconds requires keys:[].
Use a new request_id for a new action; reuse the same request_id and identical action on retry to avoid executing twice.
Only the first action from a model response can execute and return a screenshot.
Extra, stale, or invalid calls return an error without an image. Completed retries return confirmation without another image or action.
\end{tcblisting}

\clearpage
\endgroup

\end{document}